\UseRawInputEncoding
\documentclass[journal]{IEEEtran}
\usepackage{times}
\usepackage{latexsym}
\usepackage{url}
\usepackage{algorithm}
\usepackage{algcompatible}
\usepackage[noend]{algpseudocode}
\usepackage{amssymb}
\usepackage{multirow}
\usepackage{verbatim}
\usepackage{scrextend}
\usepackage{float}
\usepackage{subfig}
\usepackage{array}
\usepackage{mathtools}

\usepackage{tablefootnote}

\newcolumntype{C}[1]{>{\centering\let\newline\\\arraybackslash\hspace{0pt}}m{#1}}
\usepackage{tikz}
\usetikzlibrary{shapes,arrows}
\usepackage{amsmath,bm,times}
\usepackage{verbatim}
\usepackage{subfig}
\let\labelindent\relaxs
\usepackage{enumitem}
\usepackage{xcolor,colortbl}
\usepackage{booktabs} 
\usepackage{arydshln} 

\usepackage{float}
\usetikzlibrary{positioning}
\usepackage{assets/styles/tkz-graph}

\usetikzlibrary{decorations.pathreplacing,intersections}
\usepackage{wrapfig}
\usepackage{pbox}
\usepackage{colortbl}
\usepackage{xcolor,soul,framed}
\colorlet{shadecolor}{yellow}
\usepackage{eqparbox}
\usepackage{url}
\usepackage[square,numbers]{natbib}
\usepackage{wrapfig}
\usepackage{pifont}
\usepackage{amssymb}
\usepackage{assets/jabbrv/jabbrv}

\begin{document}

\title{A Comprehensive Understanding of Code-mixed Language Semantics using Hierarchical Transformer}

\author{Ayan Sengupta\textsuperscript{*}\thanks{\textsuperscript{*} Equal contribution.}, Tharun Suresh\textsuperscript{*}, Md Shad Akhtar, and Tanmoy Chakraborty\\ 
  \textit{Dept. of CSE, IIIT-Delhi, India} \\
  {\tt \{ayans, tharun20119, shad.akhtar, tanmoy\}@iiitd.ac.in} \\
} 

\maketitle

\begin{abstract}
Being a popular mode of text-based communication in multilingual communities, code-mixing in online social media has became an important subject to study. Learning the semantics and morphology of code-mixed language remains a key challenge, due to scarcity of data and unavailability of robust and language-invariant representation learning technique. Any morphologically-rich language can benefit from character, subword, and word-level embeddings, aiding in learning meaningful correlations. In this paper, we explore a hierarchical transformer-based architecture (HIT) to learn the semantics of code-mixed languages. HIT consists of multi-headed self-attention and outer product attention components to simultaneously comprehend the semantic and syntactic structures of code-mixed texts. We evaluate the proposed method across 6 Indian languages (Bengali, Gujarati, Hindi, Tamil, Telugu and Malayalam) and Spanish for 9 NLP tasks on 17 datasets. The HIT model outperforms state-of-the-art code-mixed representation learning and multilingual language models in all tasks. We further demonstrate the generalizability of the HIT architecture using masked language modeling-based pre-training, zero-shot learning, and transfer learning approaches. Our empirical results show that the pre-training objectives significantly improve the performance on downstream tasks.
\end{abstract}


\IEEEpeerreviewmaketitle

\section{Introduction}\label{sec:intro}
India is known for its linguistic diversity and bilingual communities. Due to such diversity, English is adopted as one of the official languages, making it ubiquitous throughout India from official purposes to the school's medium of teaching. Therefore, it is hard for the communities to avoid the influence of English in their native languages, and this results in the popular form of communication, called \textit{code-mixing}. Code-mixing ({\em aka} code-switching) is a linguistic phenomenon where two or more languages are alternatively used in conversations. This primarily makes use of a single script in case of text, most often Latin script. 
Parshad {\em et al.} \cite{PARSHAD2016375} studied the socio-linguistic aspect behind the evolution of Indian code-mixed languages and concluded the socio-economic aspect behind the adaptation of English in the Indian subcontinent. 
Given the immense popularity of this form of communication, there is a dire need to study the patterns that could better understand its linguistic properties and can be used for useful predictions. The major limitation of existing studies on code-mixed data is that the variations across alternating languages do not generalize well to all languages. This calls for an intuitive approach to identify the commonalities and differences across languages that is task-invariant and language-agnostic.

Various methodologies studied the contexts of code-mixed texts. Recent works by Pratapa {\em et al.} \cite{pratapa_word_2018} and Aguilar {\em et al.} \cite{aguilar_english_2020} presented analyses on code-mixed texts on learning meaningful representations. As most NLP tasks emphasize structural and contextual information, the former study focuses on multi-lingual embedding to understand the nuances across languages. The latter uses hierarchical attention on character $n$-grams to learn word semantics. Building on these ideas, we \cite{sengupta_hit_2021} recently explored a \textbf{HI}erarchically attentive \textbf{T}ransformer (HIT) framework that learns sub-word level representations. It employs a fused attention mechanism (FAME) -- a combination of outer product attention \cite{le_self-attentive_2020} with multi-headed self attention \cite{vaswani_attention_nodate}. 
Since code-mixed texts mostly follow informal contexts, minor misspellings tend to represent the same word with a different sub-word level representations. This is very well handled by character-level HIT that learns to represent similar words nearby in the embedding space. Finally, the character-level, sub-word-level, and word-level representations are fused to obtain a robust representation of code-mixed text. This embedding can be used to train any downstream task which requires code-mixed language processing. In this paper, we extend our earlier effort on HIT, by including extensive evaluation, new insights, and a detailed discussion on the generalization capability of HIT as the code-mixed representation learning model.

To this end, we evaluate the HIT model on 9 NLP tasks -- 4 classification tasks (sentiment classification, humour classification, sarcasm detection, and intent detection), 3 sequence labeling tasks (PoS tagging, NER, and slot-filling), and 2 generative tasks (machine translation and dialog/response generation). These tasks are spread across 6 Indian languages (Hindi, Bengali, Tamil, Telugu, Gujarati, Malayalam) and Spanish language, spanning over 17 datasets\footnote{In each case, English is the embedded language.}. Moreover, out of these tasks, 3 of them (intent detection, slot-filling, and response generation) belong to a conversational dialog setting. Our evaluation suggests that HIT learns better and robust inferences, as compared to the other state-of-the-art models. Also, the generalized word embedding of HIT can be further applicable to any downstream tasks. We show its effectiveness in representing word embedding in a contextual space and how it can be used to find similarities across inputs.

Furthermore, towards learning a task-invariant robust semantic understanding from code-mixed texts, we adopt a zero-shot learning objective to learn semantic similarity across different code-mixed texts without any explicit label. Our empirical study shows the effectiveness of zero-shot learning over traditional supervised learning objective, even for noisy code-mixed texts.

\noindent {\bf Contributions:} The contributions of the current work, in addition to our earlier work \cite{sengupta_hit_2021}, are as follows:
\begin{itemize}[leftmargin=*]
    \item We show the effectiveness of HIT on sarcasm detection, humour classification, intent detection, and slot filling tasks on code-mixed texts on 6 Indian languages.
    \item We show the effectiveness of the HIT model's word representations on generation tasks such as response generation over a conversational dataset.
    \item Our work offers a very first study on understanding the generalizability of code-mixed representation learning on downstream classification tasks.
\end{itemize}

\noindent \textbf{Reproducibility:}
We have made the source code, datasets and steps to reproduce the results public at \url{https://github.com/LCS2-IIITD/Code-mixed-classification}.

\section{Related Work}
\label{sec:lit}

Code-mixing research has been around for quite some time, and most of the work has emphasized embedding space with bilingual embedding and cross-lingual transfer as discussed in several studies \cite{upadhyay_cross-lingual_2016, ruder_survey_2019}. Akhtar {\em et al.} \cite{akhtar_solving_2018} discussed the low-resource constraints in code-mixed datasets and how bilingual word embedding can be leveraged using a parallel corpus. Extending on the parallel corpus approach, Faruqui {\em et al.} \cite{faruqui_improving_2014} proposed Canonical Correlation Analysis (CCA) to project multilingual properties in the monolingual space. Though this work has effectively helped to understand monolingual information better, the vectors encoded do not transfer well to semantic tasks as much as they do for syntactic tasks. Similar findings \cite{hermann_multilingual_2014, luong_bilingual_2015} showed learning multiple language embedding in a single embedding space.

Expanding to generation tasks which add another layer of complexity, Labutov {\em et al.} \cite{labutov_generating_2014} showed L2 method pedagogy to generate code-mixed texts. It is based on static optimization, and the generation does not perform well, bounded to the context. Gupta {\em et al.} \cite{gupta_uncovering_2018} explored the question-answering domain of the generation task. They proposed a CNN-BiGRU based model with bi-linear attention in a common embedding space that falls short of learning distinctions among the code-mixed representations. The authors also discussed a pipeline model using NER and PoS tagging for code-mixed question generation, which augments errors along the pipeline.

Banerjee {\em et al.} \cite{banerjee_dataset_2018} proposed one of the large-scale code-mixed datasets on conversation. Prior to this, there had not been any comprehensive conversational dataset in the code-mixed domain. It is built upon the DSTC2 response generation dataset, which has been code-mixed into an English-Indic language combination. This is a robust dataset built with manual annotations, which has formed the basis for response generation, intent detection, and slot filling tasks in a code-mixed setup. Srivastava {\em et al.} \cite{srivastava_phinc_2020} expanded the notions that need to be analyzed when it comes to code-mixed tasks. They discussed 6 potential issues associated with code-mixed datasets -- ambiguity, spelling variations, named-entity recognition, informal writing, misplaced punctuation, and missing context. It argued serious limitations in the current machine translation systems and hence the sequence generation tasks.

Tangentially, one of the approaches is to consider sub-word level. Since many languages in code-mixed texts are morphologically rich languages, we can leverage them in learning better representations. Prabhu {\em et al.} \cite{prabhu_towards_2016} used a CNN-LSTM model to learn sub-word embedding from 1-dimensional convolutions over character inputs. This effectively translates to better results on sentiment classification tasks on code-mixed datasets. These representations, when matched with corresponding attention mechanisms to learn the inter-dependencies, show promising results as discussed in the HAN model \cite{yang_hierarchical_2016}. Since a document represents an extended context and can involve multiple key sentences/words, the HAN model proposes hierarchical attention over the document that learns to attend to keywords and sentences, improving the classification task. Along similar lines, Aguilar {\em et al.} \cite{aguilar_english_2020} proposed CS-ELMO for code-mixed datasets, which works well using the hierarchical attention model by including the bi-gram and tri-gram level of sub-word embedding.

Language modeling is the task of learning how a text is formed. Devlin {\em et al.} \cite{devlin_bert_2019} presented a seminal work on language modeling, where the authors developed a pre-trained encoder, pre-trained on huge monolingual corpus with masked language modeling (MLM) objective. The self-supervision in MLM allows to learn representation of texts, even without using any supervision in terms of external labels. In a very recent work, Khanuja {\em et al.}~\cite{khanuja2021muril} proposed a pre-trained language model, MuRIL, that is trained on monolingual Indian texts. They explicitly augment texts with both translated and transliterated text pairs to generate parallel corpus. As compared to monolingual or bilingual language models, studies on code-mixed language model are rare. Pratapa {\em et al.} \cite{pratapa2018language} explored sampling-based language models for Hindi-English code-mixed texts. The effectiveness of pre-trained language models on downstream tasks has been shown in numerous recent works. Another popular avenue of self-supervised learning in text data is zero-shot or few-shot learning, where the representation of a text is learned on one task and is reused in other tasks. Zero-shot learning helps in understanding a better generalized representation of texts. A recent study \cite{brown2020language} showed the few-shot learning capabilities of language models. Moreover, Gupta {\em et al.} \cite{gupta2021unsupervised} adopted an unsupervised pre-training objective for the code-mixed sentiment classification task. On a similar line, Yadav {\em et al.} \cite{yadav2021zera} conducted zero-shot classification by transferring knowledge from different monolingual and cross-lingual word embeddings.

HIT \cite{sengupta_hit_2021} is one of the first efforts that take both the structural and semantic information of code-mixed texts into account, as compared to the existing one that focuses mostly on learning semantics from code-mixed data. HIT demonstrates the effectiveness of hierarchical transformer-based representation learning on 5 Indian code-mixed languages across sentiment classification, PoS, NER and machine translation tasks. Our current work focuses on the generalizability aspect of code-mixed representation learning, where we learn the representation of a code-mixed text and reuse it for multiple tasks. To the best of our knowledge, ours is the first large-scale study of code-mixed learning where we evaluate our methodology on 9 diverse tasks spanning sequence-labelling, classification, and generation tasks. We highlight the key areas where our current work extends the previous study:
\begin{itemize}[leftmargin=*]
\item This study particularly focuses on the {\em generalizibility} aspect of the model.
\item We add different pre-training objectives -- masked language modeling and zero-shot learning for better generalization and domain adaptation.
\item We extend the empirical study to more Indian languages across a variety of tasks including response generation, sarcasm detection, humour classification, intent detection and slot filling.
\end{itemize}

\section{Methodology}
\label{sec:method}

\begin{figure}[t]
\centering
\includegraphics[scale=.5]{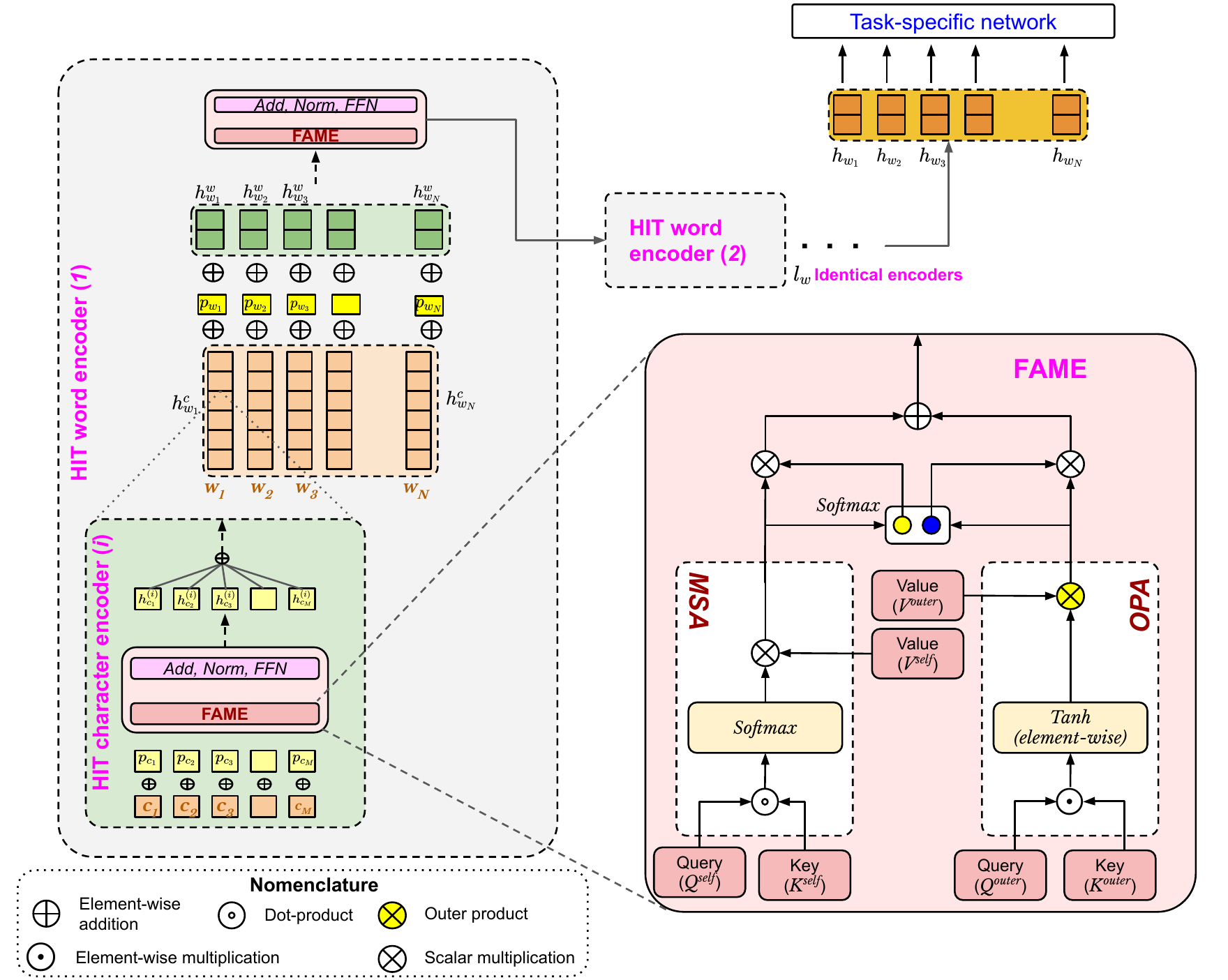}
\caption{\textbf{Hi}erarchical \textbf{T}ransformer along with our novel {\bf FAME} mechanism for attention computation.}
\label{fig:model}
\end{figure}

In this section, we describe HIT and how it incorporates character and word embeddings and hierarchical attention together to learn a robust linguistic understanding of code-mixed texts. HIT's framework is based on the encoder-decoder architecture \cite{vaswani_attention_nodate} and the hierarchical attention network \cite{yang_hierarchical_2016}. The character- and word-level HIT encoders work in a hierarchy to learn the semantics of a code-mixed sentence (interchangeably, text). Both these encoders make use of the fused attention mechanism (FAME), which is a combination of multi-headed self-attention and outer product attention \cite{le_self-attentive_2020}. The outer product attention aids in learning lower-order relationships between arbitrary pair of words and give better relational reasoning, while the multi-headed self-attention learns a higher-level semantic understanding. We illustrate the model architecture in Figure~\ref{fig:model}.

\subsection{Fused-Attention Mechanism (FAME)}

FAME is a combination of multi-headed self-attention (MSA) and outer-product attention (OPA). We extend the vanilla transformer architecture \cite{vaswani_attention_nodate} to incorporate the outer-product attention and obtain the higher-order relationships among input text. Given an input $x$, we use query, key and value weight matrices $W_Q^{self}, W_K^{self}$ and $W_V^{self}$ to project onto $Q^{self}, K^{self}$ and $V^{self}$, respectively. Likewise for OPA, we use $W_Q^{outer}, W_K^{outer}$ and $W_V^{outer}$ to obtain $Q^{outer}, K^{outer}$ and $V^{outer}$. We combine the representations learned using MSA and OPA by taking a weighted sum as follows; 
\begin{equation}
    Z = \alpha_1\cdot Z^{self} \oplus \alpha_2\cdot Z^{outer}
\end{equation}
where $\oplus$ denotes the element-wise addition while $\alpha_1$ and $\alpha_2$ are the weights learned by the softmax layer for the respective attention layers, thus producing the weighted sum output.

\begin{itemize}[leftmargin=*]
    \item \textbf{Multi-Headed Attention:} We adopt the MSA module from Vaswani {\em et al.} \cite{vaswani_attention_nodate} that makes a scaled dot product attention between query and key vectors to produce the value vector $Z^{self}$ by learning the appropriate weights as follows:
\begin{eqnarray}
Z^{self} = \sum_{i}^{N} \emph{softmax} \left( \frac{q . k_i}{\sqrt{d^k}}\right) v_i, \forall q \in Q^{self}
\end{eqnarray}
where $N$ is the length of the input sequence, and $d$ is the dimension of the key vector. 

\item \textbf{Outer-Product Attention: } We employ the outer-product attention \cite{le_self-attentive_2020} as another attention mechanism. 
The outer product attention and the multi-headed self-attention differ in terms of operators only -- their operations remain the same. OPA makes use of the row-wise \emph{tanh} activation function instead of the \emph{softmax} activation function. Additionally, MSA uses a scalar dot product, whereas OPA computes element-wise multiplication between the query and the key vectors. 
Finally, we perform outer-product between the value vector and the softmax output. As OPA helps with better relational reasoning across pairs of elements, lower-level associations are learned better. The formula is as follows:
\begin{eqnarray}
Z^{outer}  & = & \sum_{i}^{N} \emph{softmax} \left( \frac{q \odot k_i}{\sqrt{d^k}} \right) \otimes v_i, \forall q \in Q^{outer}
\end{eqnarray}
where $\odot$ is element-wise multiplication, and $\otimes$ is the outer product.
\end{itemize}

\subsection{HIT Encoders}
\begin{itemize}[leftmargin=*]
    \item \textbf{Character-level HIT:} Given a word $w_i = \{c_1, c_2, \cdots, c_m\}$ having $m$ characters, \emph{character-level} HIT leverages the formation of character sequences. The primary objective of the character-level HIT model is to understand the phonetics of code-mixed language and to bypass the need of a pre-defined word vocabulary. 
    
    The hidden representation learnt through the character-level HIT is fed to a layer-normalization layer \cite{ba2016layer} along with a residual connection. Subsequently, we pass it through a position-wise feed-forward layer. 
    In the model, we stack $l_c$ number of identical encoders, where each layer $i$ of {character-level} HIT learns a representation $\langle h^{(i)}_{c_1}, h^{(i)}_{c_2}, h^{(i)}_{c_3}, \cdots h^{(i)}_{c_m}\rangle$.
    
    Finally, we apply a hierarchical attention operator, as defined by Yang {\em et al.} \cite{yang_hierarchical_2016}, to obtain the final word representation $h^{(c)}_{w_i}$.

    \item \textbf{Word-level HIT:} We utilize the word representation obtained from character-level HIT in learning a higher order semantics for each code-mixed word. To obtain representation at the sentence level, we adapt the word-level HIT encoder to combine $\langle h^{(c)}_{w_1}, h^{(c)}_{w_2}, h^{(c)}_{w_3}, \cdots h^{(c)}_{w_n} \rangle$ with a dynamic word embedding $\langle h^{(w)}_{w_1}, h^{(w)}_{w_2}, h^{(w)}_{w_3}, \cdots h^{(w)}_{w_n} \rangle$ learned by utilizing only words. To preserve the relative positioning among different word tokens, we add the positional encoding \cite{vaswani_attention_nodate} $\langle p_{w_1}, p_{w_2}, p_{w_3}, \cdots p_{w_n} \rangle$ with the above representation. The character-level HIT encoder is shared across different word encoders. We design each encoder layer of {word-level} HIT in a similar fashion as we design for the character-level, however, considering the sequence of words as input. 
\end{itemize}

\subsection{Task-specific Layers}
We evaluate our HIT representation on various downstream tasks such as the sequence labeling, classification, and generation tasks. 
We use average pooling for the classification tasks to aggregate the word representations extracted from the word-level HIT encoder. However, for the sequence prediction tasks, we skip the average pooling and use the original word-level representation instead. In addition to the embedding learned by the HIT model, we concatenate \emph{tf-idf} based statistical feature. The {\em tf-idf} vectors capture the uni-, bi-, and tri-gram features of the inputs, which aid in eliminating handcrafted features as explained in \cite{bansal_code-switching_2020}. 
This assists in understanding the global context of the input, which, combined with hierarchical representations, yields better results. 
\section{Datasets and Tasks}
\label{sec:data_info}
In this section, we elaborate the different datasets and tasks used for evaluating our HIT framework. We report the statistics of the datasets in Table \ref{tab:dataset_all}.

\begin{itemize}[leftmargin=*]
    \item \textbf{Sentiment Classification:} We use the dataset proposed by Chakravarthi {\em et al.} \cite{chakravarthi_corpus_nodate} for Tamil and Malayalam code-mixed languages. These are the collection of comments made on YouTube videos and consist of 4 sentiment labels, namely -- \emph{positive, negative, neutral,} and \emph{mixed-feelings}. 
    For Hindi-English, we explore the code-mixed dataset for sentiment classification developed by Prabhu {\em et al.} \cite{prabhu_towards_2016}. It comprises popular public pages on Facebook. They follow a three-level polarity scale - \emph{positive, negative}, and \emph{neutral}. 
    There are about $15\%$ \emph{negative}, $50\%$ \emph{neutral}, and $35\%$ \emph{positive} comments. For the Spanglish (Spanish-English) dataset, we select the SemEval-$2020$ Task $9$ dataset \cite{patwa_semeval-2020_2020}, which is a collection of tweets collated with standard three-level polarity.
    \item \textbf{Named-Entity Recognition (NER):} For NER, we utilize Hindi \cite{singh2018named} and Spanish \cite{aguilar-etal-2018-named} datasets with $2079$ and $52781$ sentences, respectively. In Hindi, the labels are \textit{name}, \textit{location}, and \textit{organization}, while the Spanish dataset has 6 additional labels -- \textit{event}, \textit{group}, \textit{product}, \textit{time}, \textit{title}, and \textit{other}.
    
    \item \textbf{PoS Tagging:} We use $3$ different PoS datasets for Hindi-English, Bengali-English, and Telegu-English code-mixed texts. The Hindi-English code-mixed PoS dataset \cite{singh-etal-2018-twitter} has $1489$ sentences collected from Twitter. Each token in the sentence is tagged with one of the $14$ tags. The Bengali and Telugu datasets are part of the ICON-2016 workshop\footnote{\url{http://amitavadas.com/Code-Mixing.html}} and have $1982$ and $626$ sentences, respectively. These are collected from various online social network channels and contain $52$ and $39$ tags, respectively. For Spanish, we use Linguistic Code-switching Evaluation (LinCE) PoS dataset \cite{alghamdi-etal-2016-part} consisting of more that $35k$ sentences with $14$ tags. 
    
    \item \textbf{Machine Translation:} We adopt the Hindi-English code-mixed parallel corpus for machine translation \cite{gupta-etal-2020-semi} comprising more than $200k$ sentence pair.
    
    \item \textbf{Response Generation: } Built on the DSTC2 dataset \cite{henderson2014second}, Banerjee {\em et al.} \cite{banerjee_dataset_2018} made a comprehensive and faithful adaptation to Indic code-mixed languages, namely Hindi, Bengali, Gujarati, and Tamil code-mixed languages. It consists of $49k$ utterances with about $6.7k$ unique utterances. 
    The number of average utterances per dialog is $15.19$, and the vocabulary size for English dataset is $1229$. In comparison, the average levels of code-mixing in each utterance of Hindi, Bengali, Gujarati, and Tamil are $12.11$, $14.28$, $11.80$, and $12.96$, respectively.
    
    \item \textbf{Intent Detection and Slot Filling: } Further to Banerjee {\em et al.} \cite{banerjee_dataset_2018}, the intent detection and slot filling values are tagged corresponding to each utterance for all the code-mixed datasets, namely Hindi, Bengali, Gujarati, and Tamil. A total of 17 intent values are used to represent each utterance. Moreover, the dataset defines 
    7 slots to detect user request terms \emph{area}, \emph{food}, \emph{price range}, \emph{address}, \emph{postcode}, \emph{phone} and \emph{slot}. Details of intents and slot-values are depicted in Tables \ref{tab:data_intents} and \ref{tab:data_slots}, respectively. 
    
    \begin{table}[t]
\centering

\resizebox{\columnwidth}{!}{
\begin{tabular}{|l|p{5em}|l|l|l|l|l|l|}
\hline
\multicolumn{1}{|c|}{\multirow{2}{*}{\bf Tasks}} & \multicolumn{1}{c|}{\multirow{2}{*}{\bf Lang}} &
\multicolumn{2}{c|}{\bf Train} & \multicolumn{2}{c|}{\bf Test} & \multicolumn{1}{c|}{\multirow{2}{*}{\bf Total}} & \multicolumn{1}{c|}{\multirow{2}{*}{\bf \#Labels}}\\\cline{3-6}
\multicolumn{1}{|c|}{} & \multicolumn{1}{c|}{} & \multicolumn{1}{c|}{\bf \#Sent} & \multicolumn{1}{c|}{\bf \#Tokens} & \multicolumn{1}{c|}{\bf \#Sent} & \multicolumn{1}{c|}{\bf \#Tokens} & \multicolumn{1}{c|}{} & \multicolumn{1}{c|}{}\\

\hline \hline

\multirow{4}{*}{\bf POS} & Hi* & 1191 & 6,575 & 148 & 2,300 & 1,489 & 14\\
& Te* & 1,585 & 7,190 & 198 & 2,927 & 1,982 & 52\\
& Be* & 500 & 4,108 & 62 & 631 & 626 & 39\\
& Sp & 27,893 & 11,897 & 4,298 & 3,866 & 36,489 & 17\\
\hline

\multirow{2}{*}{\bf NER} & Hi* & 1,663 & 9,397 & 207 & 3,272 & 2,079 & 7\\
& Te* & 33,611 & 52,680 & 10,085 & 23,787 & 53,781 & 19\\
\hline

\multirow{4}{*}{\bf Sentiment} & Hi* & 3,103  & 9,005  & 387  & 3,191  & 3,879  & 3\\
& Ta & 11,335 & 27,476 & 3,149 & 10,339 & 15,744 & 4\\
& Ma & 4,851 & 16,551 & 1,348 & 6,028 & 6,739 & 4\\
& Sp & 12,194 & 28,274 & 1,859 & 7,822 & 15,912 & 3\\
\hline

\multirow{2}{*}{\bf MT} & En(Src) & \multirow{2}{*}{248,330} & 84,609 & \multirow{2}{*}{2,000} & 5,314 & \multirow{2}{*}{252,330} & -\\
& Hi(Tgt) & & 108,442 & & 5,797 & &\\
\hline

{\multirow{1}{*}{\bf Sarcasm/Humour}} & Hi* & 16,123  & 30,036  & 2,475  & 9,333  & 15,576  & 2\\
 \hline
 
 {\multirow{1}{*}{\bf Intent Detection}} & Be, Gu, Hi, Ta & 26,964  & 1521  & 21,386  & 580  & 48,350  & 17\\
\hline

 {\multirow{1}{*}{\bf Slot Filling}} & Be, Gu, Hi, Ta & 26,964  & 1521  & 21,386  & 580  & 48,350  & 15\\
 \hline

\multirow{4}{*}{\bf Response Generation} & Hi & 14,436 & 1576  & 2,244  & 576  & 16,680  & \multirow{4}{*}{-}\\
& Ta & 14,930 & 1600  & 2,500  & 650  & 17,430  & \\
& Gu & 14,300 & 1503  & 2,176  & 526  & 16,476  & \\
& Be & 14,100 & 1423  & 2,050  & 490  & 16,150  & \\
\hline
\end{tabular}%
}
\caption{Statistics for all datasets across all tasks. Here \textit{Total} refers to the number of samples in the entire dataset. * denotes 90-10 ratio. For intent detection and slot filling tasks, tokens are not mentioned for individual languages due to having a minimal variation among them.}
\label{tab:dataset_all}
\vspace{-2.5mm}
\end{table}
    
\begin{table*}[t]
\centering

\resizebox{\textwidth}{!}{%
\begin{tabular}{|c|c|c|c|c|c|c|c|c|c|c|c|c|c|c|c|c|c|c|}
\hline
\multicolumn{19}{|c|}{Intent Detection}\\
\hline
\multicolumn{1}{|c|}{\multirow{1}{*}{\bf Split}} &
ack & affirm & cant.help & confirm & deny & expl-conf & impl-conf & inform & negate & offer & repeat & reqalts & reqmore & request & silence & thankyou & welcomemsg &
\multicolumn{1}{c|}{\multirow{1}{*}{\bf Total}} \\\cline{3-6}

\hline \hline

\bf Train & 889 & 421 & 870 & 199 & 11 & 595 & 1009 & 8723 & 109 & 466 & 287 & 240 & 25 & 3727 & 2734 & 3070 & 1589 & 24,964\\
\bf Test & 684 & 432 & 838 & 131 & 22 & 808 & 1207 & 7144 & 82 & 337 & 743 & 284 & 165 & 2387 & 2586 & 2337 & 1199 & 21,386\\
\hline

\end{tabular}%
}
\caption{Statistics of intent values tagged with each utterance. This is common across all languages of the dataset, namely Bengali, Hindi, Malayalam and Tamil. Most of the intents are self-explanatory, some definitions to add to \textbf{expl-conf} - explicit confirmation on bot's offer, \textbf{impl-conf} - implicit confirmation on bot's offer, \textbf{reqalts} - user request to narrow down on a particular slot, \textbf{reqmore} - user request for more choices}
\vspace{-4mm}
\label{tab:data_intents}
\end{table*}
    
\begin{table}[t]
\centering

\resizebox{\columnwidth}{!}{%
\begin{tabular}{|c|c|c|c|c|c|c|c|}
\hline
Intents & Slots\\
\hline \hline
cant.help & area, food, price range, name\\
\hline
confirm, deny, expl-conf, impl-conf  & area, food, price range\\
\hline
inform & addr, area, food, name, price range, phone, postcode\\
\hline
offer & name\\
\hline
request & slot\\
\hline
\end{tabular}%
}
\caption{Mapping of slot values corresponding to each intent for which slots are tagged.}
\vspace{-5mm}
\label{tab:data_slots}
\end{table}
    
    \item \textbf{Sarcasm Detection and Humour Classification:} We use the Hindi-English code-mixed \emph{MaSaC} dataset provided by Bedi {\em et al.} \cite{bedi_multi-modal_2021}. This is collected from a popular Indian TV show \emph{`Sarabhai vs Sarabhai'}. It consists of $15,576$ utterances from 400 scenes across $50$ episodes. 
    Out of these utterances, $3,139$ are sarcastic utterances and $5,794$ are humourous utterances. A point to note is an utterance can be both sarcastic and humourous at the same time.
\end{itemize}

\noindent We obtain our final datasets through a series of data preprocessing steps described as below.

\subsection{Dataset Preprocessing}
We preprocess all the input texts by removing punctuation, hyperlinks, and address sign (@). The response generation dataset is ordered in pairs of user and system responses combined into sets of conversations with each conversation about a particular request that the user is looking for. To model an appropriate response generation task, we incorporate the history of the dialogue in each of the inputs since the conversation is around a single request and servicing the same. 
Hence, at the $j$th instance of the conversation, if the utterances of the bot are ${b_1, b_2, \cdots, b_{j-1}}$ and user inputs are ${u_1, u_2, \cdots, u_{j-1}}$, then the input given to the model to predict the $j^{th}$ bot's response $u_j$ would be like, ${b_1,u_1,b_2,u_2,\cdots b_{j-1},u_{j-1}}$. 
To handle the sentence boundaries, we add $[CLS]$ token at the start and $[EOS]$ token at the end of each input and output texts. 
For the slot filling task, we approach it as sequence prediction and create a sequence corresponding to the input (based on the slot values) in IOB format.
\section{Experiments and Results}
In this section, we elaborate on the experiments performed, results of our evaluations, and the required analyses carried out as part of the results. \vspace{-2.5mm}

\subsection{Evaluation Metrics}
For the classification and sequence-labelling tasks, we report macro Precision (Prec), Recall (Rec), and F1-scores. 
On the other hand, for the generative task, we use Rouge (\textbf{RL}) \cite{lin_rouge_nodate}, BLEU (\textbf{B}) \cite{papineni_bleu_2001}, and METEOR (\textbf{M}) \cite{banerjee_meteor_2005} for the evaluation. 

\newcommand{\nertab}{
\begin{tabular}{|l|l|l|l||l|l|l|}
\hline
\multicolumn{1}{|c|}{\multirow{2}{*}{Model}} & \multicolumn{3}{c||}{Hindi} & \multicolumn{3}{c|}{Spanish}\\ \cline{2-7} 
\multicolumn{1}{|c|}{} & \multicolumn{1}{c}{Prec} & \multicolumn{1}{c}{Rec} & \multicolumn{1}{c||}{F1} & \multicolumn{1}{c}{Prec} & \multicolumn{1}{c}{Rec} & \multicolumn{1}{c|}{F1}\\ \hline \hline

\textbf{BLSTM} & 0.622 & 0.781 & 0.579 & 0.581 & 0.659 & 0.603\\ \hline

\textbf{HAN} & 0.721 & 0.767 & 0.695 & 0.615 & \bf 0.679 & 0.644\\ \hline

\textbf{ML-BERT} & 0.792 & 0.779 & 0.714 & 0.652 & 0.623 & 0.643\\ \hline

\textbf{CS-ELMO} & 0.815 & 0.780 & 0.735 & 0.683 & 0.668 & 0.671\\ \hline

\rowcolor{lightgray}
\textbf{HIT} & \bf 0.829 & 0.788 &  \bf 0.745 &  \bf 0.695 &  0.671 &  \bf 0.684\\ \hline

\textbf{$(-) {Atn}^{outer}$} & 0.821 & 0.767 & 0.732 & 0.669 & 0.663 & 0.668\\ \hline

\textbf{$(-) char$ HIT} & 0.556 & \bf 0.815 & 0.528 & 0.498 & 0.664 & 0.539
\\ \hline
\end{tabular}%
\label{tab:ner_stats}
}

\newcommand{\sentimenttab}{
\begin{tabular}{|l|l|l|l|l|l|l|l|l|l|l|l|l|l|l|l|l|l|l|l|l|}
\hline
\multicolumn{1}{|c|}{\multirow{2}{*}{Model}} & \multicolumn{3}{c|}{Hindi} & \multicolumn{3}{c|}{Tamil}  & \multicolumn{3}{c|}{Malayalam} & \multicolumn{3}{c|}{Spanish} \\ \cline{2-13} 
\multicolumn{1}{|c|}{} & \multicolumn{1}{c}{Prec} & \multicolumn{1}{c}{Rec} & \multicolumn{1}{c|}{F1} & \multicolumn{1}{c}{Prec} & \multicolumn{1}{c}{Rec} & \multicolumn{1}{c|}{F1} & \multicolumn{1}{c}{Prec} & \multicolumn{1}{c}{Rec} & \multicolumn{1}{c|}{F1} & \multicolumn{1}{c}{Prec} & \multicolumn{1}{c}{Rec} & \multicolumn{1}{c|}{F1}\\ \hline \hline

\textbf{BLSTM-CRF} & 0.643 & 0.628 & 0.636 & 0.502 &  0.428 & 0.451 & 0.653 & 0.588 & 0.612 & 0.429 & 0.431 & 0.428  \\ \hline
\textbf{Subword-LSTM} & 0.632  & 0.634  & 0.632  & 0.503  & 0.418  & 0.426 &  0.577 &  0.592 &  0.581  & 0.445  & 0.437  & 0.432 \\ \hline
\textbf{HAN} & 0.680 & 0.671 & 0.673 & 0.490 & 0.411 & 0.439 & 0.639 & 0.611 & 0.634 & 0.449 & 0.439 & 0.440\\ \hline
\textbf{ML-BERT} & 0.609 & 0.604 & 0.599 &  0.260  & 0.310 &  0.280  & 0.600 &  \bf 0.630  & 0.610  & 0.451  & 0.419 &  0.437
\\ \hline
\textbf{CS-ELMO} & 0.679 &  0.661 &  0.667 &  0.515 &  0.432  & 0.459 &  0.666  & 0.623 &  0.642 &  0.429  & 0.453 &  0.431
\\ \hline
\rowcolor{lightgray}
\textbf{HIT} & \bf 0.745 &  \bf 0.702 &  \bf 0.703 &  0.499 &  \bf 0.451 &  \bf 0.473  & 0.710 &  0.628  & 0.651  & \bf 0.502  & \bf 0.454  & \bf 0.460
\\ \hline
\textbf{$(-) {Atn}^{outer}$} & 0.687  & 0.665  & 0.667  & \bf 0.520  & 0.448  & 0.455 &  \bf 0.718  & 0.624  & \bf 0.655  & 0.463  & 0.440  & 0.445
\\ \hline
\textbf{$(-) char$ HIT} & 0.652  & 0.631  & 0.650  & 0.504 &  0.418 &  0.432  & 0.659  & 0.605  & 0.627 &  0.448  & 0.438  & 0.433
\\ \hline
\end{tabular}%
\label{tab:sentiment_stats}
}

\newcommand{\postab}{
\begin{tabular}{|l|l|l|l|l|l|l|l|l|l|l|l|l|l|l|l|l|l|l|l|l|}
\hline
\multicolumn{1}{|c|}{\multirow{2}{*}{Model}} & \multicolumn{3}{c|}{Hindi} & \multicolumn{3}{c|}{Telugu}  & \multicolumn{3}{c|}{Bengali} & \multicolumn{3}{c|}{Spanish} \\ \cline{2-13} 
\multicolumn{1}{|c|}{} & \multicolumn{1}{c}{Prec} & \multicolumn{1}{c}{Rec} & \multicolumn{1}{c|}{F1} & \multicolumn{1}{c}{Prec} & \multicolumn{1}{c}{Rec} & \multicolumn{1}{c|}{F1} & \multicolumn{1}{c}{Prec} & \multicolumn{1}{c}{Rec} & \multicolumn{1}{c|}{F1} & \multicolumn{1}{c}{Prec} & \multicolumn{1}{c}{Rec} & \multicolumn{1}{c|}{F1}\\ \hline \hline
\textbf{BLSTM-CRF} & 0.821 & 0.913 & 0.782 & 0.595 & 0.747 & 0.572 & 0.842 & 0.851 & 0.817 & 0.704 & 0.836 & 0.680  \\ \hline
\textbf{HAN} & 0.802 & 0.879 & 0.815 & 0.693 & 0.701 & 0.684 & 0.811 & 0.823 & 0.818 & 0.497 & 0.629 & 0.527\\ \hline
\textbf{ML-BERT} & 0.833 & 0.884 & 0.847 & 0.802 & 0.762 & 0.771 & 0.793 & 0.815 & 0.807 & 0.853 & 0.808 & 0.802
\\ \hline
\textbf{CS-ELMO} & 0.885 & \bf 0.961 & 0.910 & 0.831 & 0.790 & 0.775 &  \bf 0.873 & 0.851 & 0.847 & 0.740 & \bf 0.835 & 0.729
\\ \hline
\rowcolor{lightgray}
\textbf{HIT} & \bf 0.918 & 0.955 & \bf 0.919 & 0.815 & 0.749 & 0.762 & 0.841 & \bf 0.855 & \bf 0.853 & \bf 0.871 & 0.822 & \bf 0.825
\\ \hline
\textbf{$(-) {Atn}^{outer}$} & 0.893 & 0.948 & 0.914 & \bf 0.839 & \bf 0.793 & \bf 0.786 & 0.839 & 0.852 & 0.845 & 0.859 & 0.813 & 0.820
\\ \hline
\textbf{$(-) char$ HIT} & 0.686 & 0.922 & 0.708 & 0.629 & 0.758 & 0.626 & 0.802 & 0.830 & 0.819 & 0.723 & 0.796 & 0.732
\\ \hline
\end{tabular}%
\label{tab:pos_stats}
}

\newcommand{\mttab}{
\begin{tabular}{|l|l|l|l|l|l}
\hline
\multicolumn{1}{|c|}{\multirow{1}{*}{Model}} & \multicolumn{1}{c}{B} & \multicolumn{1}{c}{RL} & \multicolumn{1}{c|}{M}\\ \hline \hline

\textbf{Seq2Seq†} & 15.49 & 35.29 & 23.72\\ \hline

\textbf{Attentive-Seq2Seq†} & 16.55 & 36.25 & 24.97\\ \hline

\textbf{Pointer Generator†} & 17.62 & 37.32 & 25.61\\ \hline

\textbf{GFF-Pointer†} & 21.55 & 40.21 & 28.37\\ \hline


\rowcolor{lightgray}
\textbf{HIT} &  \bf 28.22 &  \bf 51.52 &  \bf 29.59\\ \hline

\textbf{$(-) {Atn}^{outer}$} & 25.95 & 49.19 & 27.63
\\ \hline
\textbf{$(-) char$ HIT} & 21.83 & 42.19 & 27.89\\ \hline
\end{tabular}
\label{tab:nmt_stats}
}


  


\begin{table*}[t]
  \centering
  \subfloat[Sentiment classification]{\scalebox{0.7}\sentimenttab}%
  \quad
  \subfloat[Named Entity Recognition]{\scalebox{0.75}\nertab}
  
  \subfloat[PoS tagging]{\scalebox{0.8}\postab}%
  \quad
  \subfloat[Hindi-English MT]{\scalebox{0.85}\mttab}
  \caption{Experimental results. Highest scores are highlighted in bold.}
  \label{tab:sentner}%
  \vspace{-2.5mm}
\end{table*}

\subsection{Baseline Models}
\begin{itemize}[leftmargin=*]
    \item{\textbf{BiLSTM}} \cite{hochreiter_long_1997}: It is presented as a preliminary neural network baseline. 
    For competitive predictions on sequence tasks, Conditional Random Fields (CRF) layer on the top is incorporated for the final classification. 
    
    \item{\textbf{Subword-LSTM}} \cite{prabhu_towards_2016}: Instead of word-level or character-level representations, Prabhu et al. \cite{prabhu_towards_2016} proposed to learn the subword-level representations as a linguistic prior. Following a convolutional and a max pool layer, it applies 2 LSTM layers.  
    We restrict ourselves to using this model only for \emph{sentiment classification} as the model does not take word boundaries into account.
    
    \item{\textbf{ML-BERT}} \cite{devlin_bert_2019}: It is trained on BERT \cite{devlin_bert_2019} to handle multilingual tasks. BERT uses the transformer architecture effectively by using the pre-training objective, called Masked Language Modeling (MLM).
    
    \item{\textbf{HAN}} \cite{yang_hierarchical_2016}: Hierarchical Attention network (HAN) combines 2 encoders, one at the word level and another at the sentence level, with attention for each level of the hierarchy.
    
    \item{\textbf{CS-ELMo}} \cite{aguilar_english_2020}: It is based on ELMo \cite{peters_deep_2018}, which is trained on English language and uses transfer learning to extend to other code-mixed languages. It is one of the the most recent and state-of-the-art models for code-mixed languages. It has shown to leverage word representations as a function of the entire input sequence giving the model a distinctive advantage.
    
    \item{\textbf{MURIL}} \cite{khanuja2021muril}: MURIL is a recently developed language model pretrained over large corpora of Indian languages. It uses monolingual corpora of 17 Indian languages translated and transliterated for pre-training purposes and reports the state-of-the-art results on the cross-lingual XTREME benchmark~\cite{hu2020xtreme}.
\end{itemize}
\subsection{Experimental Setup}
For all tasks, we consider categorical cross entropy (CE) as the loss function. We employ Adam optimizer \cite{kingma_adam_2017} with $\eta = 0.001, \beta_1 = 0.9 , \beta_2 = 0.999$ and train for $500$ \emph{epochs}. We use $dropout = 0.2$ for the regularization, whereas \emph{batch size} is set to $32$. Further, to dynamically reduce learning rate on plateaus, we monitor the validation loss with a \emph{patience} of $20$ epochs and reduce the learning rate by a factor of $0.7$. All the models are trained with early stopping ($patience = 100$) to reduce over-fitting. For ML-BERT and MURIL models, we use a smaller learning rate of 2e-5. These two models are adopted from HuggingFace Transformers library\footnote{https://github.com/huggingface/transformers}. We extract an $128$-dimensional vector for each word and character token. The maximum text length used is $40$ and maximum word length is set as $20$. To extract the statistical features from the input texts, we use TfIdfVectorizer\footnote{https://scikit-learn.org/stable/modules/generated/sklearn.feature\_extraction\\.text.TfidfVectorizer.html} from Sklearn library. We concatenate two different tfidf vectors - using word features (after removing stopwords) and the other with character features. In both these methods, we use $n$-$gram$ features with $n \in \{1,2,3\}$. To remove too frequent and rare tokens, the minimum frequency of token in the corpus is set to $2$ and the maximum frequency is set to $6$. 

\subsection{Experimental Results}
We explore variations of the features fed into the model. Precisely, we include ablations on our model on 2 of the modules. They are termed as \emph{$(-) Atn^{outer}$} and \emph{$(-)$char HIT} in all the result tables. The former's experiments are carried out by removing the outer attention module, which we have essentially fused with existing self-attention module, an important component of our FAME architecture. The latter is achieved by excluding character-level embeddings from HIT Transformer and studies the effects of it. Both these ablations are performed across all tasks. We evaluate HIT and all other baselines for all the tasks in subsequent sections. We elaborate the comparison between HIT and MURIL in a separate subsection. 

{\bf Sentiment Classification:} We show results in Table \ref{tab:sentiment_stats}. On comparison, we observe that HIT outperforms the baselines in all languages on the basis of F1 scores. For Hindi, CS-ELMo and HAN perform well; HIT considerably outperforms them by $3.6\%$.  
We observe similar phenomena for all languages with HIT reporting a minimum of $2\%$ improvement over the best baseline (HAN) -- for \emph{Spanish}, HAN is the better performing baseline among all.
The ablation study is shown in Table \ref{tab:sentiment_stats}. We see that removing character-level embeddings from the input features has a detrimental effect across the languages in comparison with all the metrics. On the other hand, except for Malayalam, removing outer-product attention reduces the performance of HIT model. In all, HIT produces state-of-the-art results.

{\bf Named-Entity Recognition:} We show results in Table \ref{tab:ner_stats}. As it is observed in the previous task, HIT outperforms the existing systems on both Hindi and Spanish languages in NER. Likewise, CS-ELMo is the better performing baseline among all; however, HIT reports $1.2\%$ better F1-score for both Hindi-English and Spanish-English. This conveys that the HIT model translates well for sequence tagging tasks as well.

{\bf PoS Tagging:} We present results of PoS tagging for different languages in Table \ref{tab:pos_stats}. We observe that HIT consistently performs better than the baselines for majority of the languages -- it outperforms as high as $2.3\%$ for Spanish based on F1 metric. On the other hand, CS-ELMo performs better among the baselines across three languages. For Spanish, ML-BERT performs closest to our HIT model. HIT ablations in Table \ref{tab:pos_stats} show that the importance of subword-level representation learning as their absence drops the performance by over $14\%$ on average across all the datasets.

\newcommand{\sartab}{
\begin{tabular}{|l|l|l|l|l|l}
\hline
\multicolumn{1}{|c|}{\multirow{2}{*}{Model}} & \multicolumn{3}{c|}{Hindi}\\ \cline{2-4} 
\multicolumn{1}{|c|}{} & \multicolumn{1}{c}{Prec} & \multicolumn{1}{c}{Rec} & \multicolumn{1}{c|}{F1}\\ \hline \hline

\textbf{BLSTM} & 0.480 & 0.482 & 0.473\\ \hline

\textbf{HAN} & 0.492 & 0.495 & 0.487\\ \hline

\textbf{ML-BERT} & 0.504 & 0.501 & 0.484\\ \hline

\textbf{CS-ELMO} & 0.494 &  0.496 & 0.489\\ \hline


\rowcolor{lightgray}
\textbf{HIT} & 0.478 &  0.487 &  0.475\\ \hline

\textbf{$(-) {Atn}^{outer}$} & \bf 0.505 & \bf 0.502 & \bf 0.490\\ \hline

\textbf{$(-) char$ HIT} & 0.480  & 0.489  & 0.479
\\ \hline
\end{tabular}%
\label{tab:sarcasm_stats}
}

\newcommand{\humtab}{
\begin{tabular}{|l|l|l|l|l|l}
\hline
\multicolumn{1}{|c|}{\multirow{2}{*}{Model}} & \multicolumn{3}{c|}{Hindi}\\ \cline{2-4} 
\multicolumn{1}{|c|}{} & \multicolumn{1}{c}{Prec} & \multicolumn{1}{c}{Rec} & \multicolumn{1}{c|}{F1}\\ \hline \hline

\textbf{BLSTM} & 0.592 & 0.576 & 0.578\\ \hline

\textbf{HAN} & 0.588 & 0.580 & 0.580\\ \hline

\textbf{ML-BERT} & 0.565 & 0.569 & 0.567\\ \hline

\textbf{CS-ELMO} & 0.573 &  0.570 &  0.570\\ \hline


\rowcolor{lightgray}
\textbf{HIT} & \bf 0.592 & \bf 0.596 & \bf 0.593\\ \hline

\textbf{$(-) {Atn}^{outer}$} & 0.590 &  0.593 &  0.591\\ \hline

\textbf{$(-) char$ HIT} & 0.583  & 0.586  & 0.584
\\ \hline
\end{tabular}
\label{tab:humour_stats}
}

\begin{table}%
  \centering
  \subfloat[Sarcasm detection]{\scalebox{0.75}\sartab}%
  \qquad
  \subfloat[Humour classification]{\scalebox{0.75}\humtab}
  \caption{Experimental results on the \emph{MaSaC} dataset.}%
  \label{tab:sarhum}%
  \vspace{-5mm}
\end{table}

\newcommand{\slot}{
\begin{tabular}{|l|l|l|l|l|l|l|l|l|l|l|l|l|l|l|l|l|l|l|l|l|l|l|l|}
\hline
\multicolumn{1}{|c|}{\multirow{2}{*}{Model}} & \multicolumn{3}{c|}{Hindi} & \multicolumn{3}{c|}{Tamil} & \multicolumn{3}{c|}{Bengali} & \multicolumn{3}{c|}{Gujarati} \\ \cline{2-16} 
\multicolumn{1}{|c|}{} & \multicolumn{1}{c}{Prec} & \multicolumn{1}{c}{Rec} & \multicolumn{1}{c|}{F1} & \multicolumn{1}{c}{Prec} & \multicolumn{1}{c}{Rec} & \multicolumn{1}{c|}{F1} & \multicolumn{1}{c}{Prec} & \multicolumn{1}{c}{Rec} & \multicolumn{1}{c|}{F1} & \multicolumn{1}{c}{Prec} & \multicolumn{1}{c}{Rec} & \multicolumn{1}{c|}{F1}\\ \hline \hline
\textbf{BLSTM-CRF} & 0.924 & 0.919 & 0.917 & 0.911 & 0.912 & 0.911  & 0.927 & 0.928 & 0.927 & 0.930 & 0.907 & 0.915  \\ \hline
\textbf{HAN} & 0.930 & 0.921 & 0.924 & \bf 0.915 & 0.921 & 0.921 & 0.931 & 0.931 & 0.929 & 0.942 & \bf 0.936 & \bf 0.938 
\\ \hline
\textbf{ML-BERT} & 0.927 & 0.922 & 0.924 & 0.913 & 0.921 & \bf 0.924 & 0.932 & 0.929 & 0.928 & 0.933 & 0.909 & 0.917  
\\ \hline
\textbf{CS-ELMO} & \bf 0.936 & \bf 0.925 & \bf 0.930 & 0.913 & 0.921 & 0.922 & 0.931 & 0.928 & 0.928 & 0.935 & 0.935 & 0.934   \\ \hline
\rowcolor{lightgray}
\textbf{HIT} & 0.913 & 0.924 & 0.918 & \bf 0.915 & \bf 0.923 & \bf 0.924 & \bf 0.949 & \bf 0.937 & \bf 0.942 & \bf 0.944 & 0.913 & 0.925 
\\ \hline
\textbf{$(-) {Atn}^{outer}$} & 0.928 & 0.921 & 0.923 & 0.910 & 0.919 & 0.921 & 0.927 & 0.929 & 0.926 & 0.931 & 0.909 & 0.916 
\\ \hline
\textbf{$(-) char$ HIT}  & 0.915 & 0.921 & 0.917 & 0.913 & 0.919 & 0.923 & 0.934 & 0.927 & 0.928 & 0.938 & 0.893 & 0.910
\\ \hline
\end{tabular}%
\label{tab:slot_filling}
}

\newcommand{\intent}{
\begin{tabular}{|l|l|l|l|l|l|l|l|l|l|l|l|l|l|l|l|l|l|l|l|l|l|l|l|}
\hline
\multicolumn{1}{|c|}{\multirow{2}{*}{Model}} & \multicolumn{3}{c|}{Hindi} & \multicolumn{3}{c|}{Tamil}  & \multicolumn{3}{c|}{Bengali}  & \multicolumn{3}{c|}{Gujarati} \\ \cline{2-16} 
\multicolumn{1}{|c|}{} & \multicolumn{1}{c}{Prec} & \multicolumn{1}{c}{Rec} & \multicolumn{1}{c|}{F1} & \multicolumn{1}{c}{Prec} & \multicolumn{1}{c}{Rec} & \multicolumn{1}{c|}{F1} & \multicolumn{1}{c}{Prec} & \multicolumn{1}{c}{Rec} & \multicolumn{1}{c|}{F1} & \multicolumn{1}{c}{Prec} & \multicolumn{1}{c}{Rec} & \multicolumn{1}{c|}{F1}\\ \hline \hline
\textbf{BLSTM} & 0.901 & 0.885 & 0.865 & 0.911 & 0.908 & 0.904 & 0.908 & 0.899 & 0.894 & 0.931 & 0.922 & 0.915  \\ \hline
\textbf{HAN}  & 0.910 & 0.892 & 0.880 & \bf 0.932 & 0.927 & \bf 0.921 & 0.916 & 0.908 & 0.897 & 0.932 & 0.924 & 0.917
\\ \hline
\textbf{ML-BERT} & 0.911 & 0.901 & 0.883 & 0.925 & 0.918 & 0.909 & 0.919 & 0.911 & 0.899 & 0.908 & 0.887 & 0.868 
\\ \hline
\textbf{CS-ELMO} & 0.906 & 0.894 & 0.882 & 0.924 & 0.924  & 0.915 & 0.927 &  0.923 &  0.916 & 0.909 & 0.900 & 0.885
\\ \hline
\rowcolor{lightgray}
\textbf{HIT} & 0.906 & \bf 0.908 & \bf 0.893 & 0.918 & 0.917 & 0.907 & \bf 0.935 &  \bf 0.929 & \bf 0.920 & \bf 0.947 & \bf 0.926 & \bf 0.928 
\\ \hline
\textbf{$(-) {Atn}^{outer}$} & \bf 0.913 & 0.895 & 0.880 & 0.927 & \bf 0.929 & \bf 0.921 & 0.919  & 0.880  & 0.863 & 0.906 & 0.899 & 0.886 
\\ \hline
\textbf{$(-) char$ HIT} & 0.900 & 0.881 & 0.862 & 0.883 & 0.861 & 0.839 & 0.900  & 0.888  & 0.873 & 0.906 & 0.905 & 0.892 
\\ \hline
\end{tabular}%
\label{tab:slot_intents}
}

\begin{table*}[t]
  \centering
  \subfloat[Intent detection]{\resizebox{0.5\textwidth}{!}{\intent}}%
  \subfloat[Slot filling]{\resizebox{0.5\textwidth}{!}{\slot}}
  \caption{Experimental results on the intent detection and slot filling tasks.}%
  \label{tab:slot_intent_table}%
\end{table*}

\begin{table*}[t]
\centering
\resizebox{0.58\textwidth}{!}{%
\begin{tabular}{|l|l|l|l|l|l|l|l|l|l|l|l|l|l|l|l|l|l|l|l|l|l|l|l|l|}
\hline
\multicolumn{1}{|c|}{\multirow{2}{*}{Model}} & \multicolumn{3}{c|}{Hindi} & \multicolumn{3}{c|}{Tamil} & \multicolumn{3}{c|}{Bengali}  & \multicolumn{3}{c|}{Gujarati}\\ \cline{2-16} 
\multicolumn{1}{|c|}{} & \multicolumn{1}{c}{B} & \multicolumn{1}{c}{RL} & \multicolumn{1}{c|}{M} & \multicolumn{1}{c}{B} & \multicolumn{1}{c}{RL} & \multicolumn{1}{c|}{M} & \multicolumn{1}{c}{B} & \multicolumn{1}{c}{RL} & \multicolumn{1}{c|}{M} & \multicolumn{1}{c}{B} & \multicolumn{1}{c}{RL} & \multicolumn{1}{c|}{M}\\ \hline \hline

\textbf{BLSTM} & 26.66 & 39.12 & 10.62 & 30.61 & 40.85 & 13.04 & 23.50 & 31.47 & 7.11 &  22.11 & 38.33 & 10.84 \\ \hline
\textbf{HAN} & 30.24 & 41.62 & 11.42 & 29.91 & 44.85 & 12.18 & 28.50 & 44.47 & 9.37 &  \bf 29.22 & 41.73 & 11.21 
\\ \hline
\textbf{ML-BERT} & 21.16 & 31.51 & 10.11 & 25.55 & 26.66 & 8.02 & 22.22 & 30.02 & 5.94 & 19.22 & 36.06 & 9.83 
\\ \hline
\textbf{CS-ELMO} & 22.83 & 36.69 & \bf 12.40 & 28.77 & 28.77 &  8.18 & 23.64  & 32.93 & 6.29 & 18.05 & 35.96 & 10.62 
\\ \hline
\rowcolor{lightgray}
\textbf{HIT} & 27.70 & 44.21 & 10.77 & 32.79 & 46.81 &  \bf 14.26 & 31.55 & \bf 47.71 & 10.15 & 27.49 & \bf 49.29 &  \bf 12.43 
\\ \hline
\textbf{$(-) {Atn}^{outer}$} &  \bf 32.70 &  \bf 45.96 & 10.22 & 34.16 &  \bf 48.21 & 13.47 & 33.06 &	46.16 &  \bf 10.50 & 23.79 & 43.53 & 11.69 
\\ \hline
\textbf{$(-) char$ HIT} & 32.25 & 44.74 & 10.07 & \bf  35.42 & 45.77 & 8.85 & \bf 33.41 & 46.79 & 8.24 & 22.15 & 43.30 & 8.38 
\\ \hline
\end{tabular}%
}
\caption{Experimental results on response generation.}
\label{tab:response_prediction}
\vspace{-5mm}
\end{table*}

{\bf Machine Translation:} Results for the machine translation task is compiled in Table \ref{tab:nmt_stats}. We compare HIT with the following baselines: {Seq2Seq} \cite{sutskever_sequence_nodate}, {Attentive-Seq2Seq} \cite{bahdanau_neural_2016}, {Pointer-Generator} \cite{see_get_2017} and {GFF-Pointer} \cite{gupta-etal-2020-semi}. We observe that HIT considerably outperforms all baselines across all metrics -- especially on ROUGE-L with a difference of 9 points compared to the best baseline. We also observe the significant effect of char-level encoding in HIT -- we obtain 21.83 BLEU, 42.19 Rouge, and 27.89 METEOR scores without the char-level encoding against 28.22 BLEU, 51.52 Rouge, and 29.59 METEOR scores with char-level encoding. 

{\bf Sarcasm Detection and Humour Classification:} We present results for the sarcasm detection and humour classification tasks in Table \ref{tab:sarhum}. We observe that HIT reports the best results against all baselines in the humour detection task -- it yields 0.593 F1-score as compared to the best baseline (HAN) F1-score of 0.580. We also observe that removing outer-product attention or the char-level encoding results in inferior performance. On the other hand, HIT obtains a comparable F1-score of 0.475 against the best F1-score of 0.490 without the outer-product attention module in the sarcasm detection task. Moreover, the performances of other baselines are also superior to the HIT's performance.
{\bf Intent Detection and Slot Filling:} We compile the experimental results for the intent detection and slot-filling tasks in Tables \ref{tab:slot_intents} and \ref{tab:slot_filling}, respectively. The HIT model outperforms all baselines in all the languages. For the Hindi, Bengali, and Gujarati languages, HIT obtains improvements of +0.011, +0.004, and +0.011 points in F1-scores, respectively, against the best baseline. Among baselines, both HAN and CS-ELMO are better in 2 languages each -- HAN in Gujarati and Tamil; CS-ELMO is Hindi and Bengali.

In the slot-filling task, HIT achieves better performance than all other baselines in 2 out of 4 languages. Among all languages, the margin is significantly higher in Bengali, where HIT achieves $1.3\%$ better F1-score than HAN, the best baseline. For Tamil, both HIT and ML-BERT achieve $0.924$ F1 score, albeit HIT achieves better precision than ML-BERT. On the other hand, HAN performs the best in Gujarati with an F1-score of $0.938$. In comparison, omitting outer-product attention in HIT yields better score than vanilla HIT in Hindi. Moreover, CS-ELMO turns out to be the best baseline for Hindi, achieving $0.012$ points better F1 than HIT.
{\bf Response Generation:} The results for the response generation tasks in 4 languages -- Hindi, Tamil, Bengali, and Gujarati -- are reported in Table~\ref{tab:response_prediction}. Similar to the machine translation task, we employ BLEU, ROUGE, and METEOR scores to evaluate the performance of the generated response. 

We observe that HIT and its variants report better scores for all three metrics in majority of the cases -- except for the 2 cases where CS-ELMO and HAN obtain better METEOR and BLEU scores in Hindi and Gujarati, respectively. Moreover, with HIT, we obtain the best scores for 4 out 12 cases (3 each for 4 languages) in the range of 1-8 improvement points against comparative baselines. In other cases, HIT yields comparative results against its variants. 

In particular, we note that the HIT model without character embeddings performs better than original HIT for majority of the languages. We hypothesize that in generative tasks like response generation, getting rid of the sub-word level representations might aid in reducing noises from input sequence and assist in achieving better generative performance. \vspace{-0.1mm}

\newcommand{\classification}{
    \begin{tabular}{|l|c|ccc|ccc|ccc|c|}
        \hline 
         \multirow{2}{*}{Task} & \multirow{2}{*}{Lang} & \multicolumn{3}{c|}{Baseline*} & \multicolumn{3}{c|}{MURIL} & \multicolumn{3}{c|}{HIT}  \\ \cline{3-11}
         & & P & R & F & P & R & F & P & R & F \\ \hline \hline
         \multirow{4}{*}{Sent} & \bf Hi & 0.680 & 0.671 & 0.673 & 0.661 & 0.678 & 0.669 & \bf 0.745 & \bf 0.702 & \bf 0.703  \\ \cdashline{2-11}
         & \bf Tm & 0.515 & 0.432 & 0.459 &  0.444 & 0.408 & 0.425 & \bf 0.520 & \bf 0.451 & \bf 0.473 \\ \cdashline{2-11}
         & \bf Mm & 0.666 & 0.630 & 0.642 & 0.621 & \bf 0.643 & 0.630 & \bf 0.718 & 0.628 & \bf 0.655 \\ \cdashline{2-11}
         & \bf Sp & 0.451 & 0.453 & 0.440 & 0.429 & 0.423 & 0.424 & \bf 0.502 & \bf 0.454 & \bf 0.460 \\ \hline
         \multirow{4}{*}{PoS} & \bf Hi & 0.885 & 0.961 & 0.910 & 0.904 & \bf 0.974 & 0.914 & \bf 0.918 & 0.955 & \bf 0.919 \\ \cdashline{2-11}
         & \bf Te & 0.831 & 0.790 & 0.775 & \bf 0.844 & \bf 0.798 & \bf 0.786 & 0.839 & 0.793 & \bf 0.786 \\ \cdashline{2-11}
         & \bf Bn & \bf 0.873 & 0.851 & 0.847 & \bf 0.873 & 0.824 & 0.846 & 0.841 & \bf 0.855 & \bf 0.853\\ \cdashline{2-11}
         & \bf Sp & 0.853 & \bf 0.835 & 0.802 & 0.862 & 0.824 & 0.809 & \bf 0.871 & 0.822 & \bf 0.825\\ \hline
         \multirow{2}{*}{NER} & \bf Hi & 0.815 & 0.780 & 0.735 & 0.817 & 0\bf .821 & 0.739 & \bf 0.829 & 0.815 & \bf 0.745\\ \cdashline{2-11}
         & \bf Sp & 0.683 & 0.679 & 0.671 & 0.659 & 0.664 & 0.670 & \bf 0.695 & \bf 0.671 & \bf 0.684 \\ \hline
         
         \multirow{1}{*}{Sarcasm} & \bf Hi & 0.504 & 0.501 & 0.489 & \bf 0.565 & \bf 0.541 & \bf 0.551 & 0.505 & 0.502 & 0.490\\ \hline
         \multirow{1}{*}{Humour} & \bf Hi & 0.592 & 0.580 & 0.580 & \bf 0.622 & \bf 0.625 & \bf 0.614 & 0.592 & 0.596 & 0.593\\ \hline
         \multirow{4}{*}{Intent} & \bf Hi & 0.911 & 0.901 & 0.883 & 0.912 & \bf 0.911 & \bf 0.905 &\bf 0.913 & 0.908 & 0.893\\ \cdashline{2-11}
         & \bf Tm & \bf 0.932 & 0.927 & \bf 0.921 & 0.918 & 0.917 & 0.907 & 0.927 & \bf 0.929 & \bf 0.921\\ \cdashline{2-11}
         & \bf Bn & 0.927 & 0.923 & 0.916 & \bf 0.941 & \bf 0.935 & \bf 0.931 & 0.935 & 0.929 & 0.920\\ \cdashline{2-11}
         & \bf Gj & 0.931 & 0.924 & 0.917 & 0.941 & 0.925 & 0.927 & \bf 0.947 & \bf 0.926 & \bf 0.928\\ \hline
         \multirow{4}{*}{Slot-fill} & \bf Hi & 0.936 & 0.925 & 0.930 & \bf 0.944 & \bf 0.943 & \bf 0.943 & 0.928 & 0.924 & 0.923\\ \cdashline{2-11}
         & \bf Tm & 0.915 & 0.921 & \bf 0.924 & \bf 0.918 & 0.914 & 0.914 & 0.915 & \bf 0.923 & \bf 0.924\\ \cdashline{2-11}
         & \bf Bn & 0.932 & 0.931 & 0.929 & 0.948 & \bf 0.947 & \bf 0.948 & \bf 0.949 & 0.937 & 0.942\\ \cdashline{2-11}
         & \bf Gj & 0.942 & \bf 0.936 & \bf 0.938 & 0.932 & 0.927 & 0.928 & \bf 0.944 & 0.913 & 0.925\\ \hline
    \end{tabular}
    \label{tab:muril_classification}
}
    
\newcommand{\generation}{
    \begin{tabular}{|l|c|ccc|ccc|ccc|c|}
        \hline 
         \multirow{2}{*}{Task} & \multirow{2}{*}{Lang} & \multicolumn{3}{c|}{Baseline} & \multicolumn{3}{c|}{MURIL} & \multicolumn{3}{c|}{HIT}  \\ \cline{3-11}
         & & B & R & M & B & R & M & B & R & M \\ \hline \hline
        \multirow{1}{*}{MT} & \bf Hi-En & 21.55 & 40.21 & 28.37 & 27.82 & \bf 54.28 & 29.06 & \bf 28.22 & 51.52 & \bf 29.58\\ \hline
        \multirow{4}{*}{Res. Gen} & \bf Hi & 30.24 & 41.62 & 12.40 & \bf 36.05 & \bf 52.18 & \bf 16.08 & 32.70 & 45.96 & 10.77\\ \cdashline{2-11}
        & \bf Tm & 30.61 & 44.85 & 13.04 & 30.46 & 37.63 & 10.27 & \bf 35.42 & \bf 48.21 & \bf 14.26\\ \cdashline{2-11}
         & \bf Bn & 28.50 & 44.47 & 9.37 & 31.08 & 46.14 & 9.70 & \bf 33.41 & \bf 47.71 & \bf 10.50\\ \cdashline{2-11}
         & \bf Gj & 29.22 & 41.73 & 11.21 & 25.38 & 43.15 & 11.69 & \bf 27.49 & \bf 49.29 & \bf 12.43\\ \hline
    \end{tabular}
    \label{tab:muril_generation}
    }

\begin{table}[t]
  \centering
  \subfloat[Classification and Sequence labelling tasks]{\scalebox{0.7}\classification}%
  \quad
  \subfloat[Generation Tasks]{\scalebox{0.7}\generation}
  \caption{Comparative study between HIT and MURIL. * denotes the respective best baseline system from Tables \ref{tab:sentner} -- \ref{tab:response_prediction}.} 
  \label{tab:muril_comparison}%
  \vspace{-5mm}
\end{table}

\subsection{Comparison with MURIL}
Recently, Khanuja et al. \cite{khanuja2021muril} proposed MURIL, a large-scale pre-trained language model for Indian languages. It reports state-of-the-art performances for multiple tasks. In this section, we compare HIT with MURIL elaborately and understand the strengths and weaknesses of these methods. We report the performances of MURIL and HIT along with the best performing baselines in respective tasks in Table~\ref{tab:muril_comparison}.
In the sentiment classification task (c.f. Table~\ref{tab:muril_classification}), we observe that HIT outperforms MURIL in all 4 languages, with a wide margin of $3\%$ F1-score. We further observe that the MURIL's performance is inferior to the best baseline as well. We argue that the superior performances of HIT and the best baselines (\textit{viz.} CS-ELMO and HAN) against MURIL is due the effectiveness of the sub-word and hierarchical representations for learning semantics in code-mixed texts.

Even on PoS tagging, we observe a similar trend, in which HIT achieves $1\%$ better F1 score on average, as compared to MURIL, across all the languages. The difference is wider for Spanish, possibly due of the pre-training objective of MURIL. Similarly, for the NER classification task, supervised representation learning methods like HIT and CS-ELMo perform better than MURIL. On the other hand, in the sarcasm detection and humour classification tasks, we observe that MURIL outperforms (by $6\%$ and $2\%$, respectively) HIT. This could be attributed to the fact that the \textit{MaSaC} dataset contains over $32\%$ of monolingual Hindi text transliterated to English, which goes in favor of MURIL. Similarly, in the intent classification and slot-filling tasks, MURIL tends to perform better than HIT and other baselines. We observe that $40\%$ of the tokens in these datasets are either in English or language invariant, which aids in superior performance of MURIL. 
We report the comparison of HIT and MURIL on generative tasks in Table~\ref{tab:muril_generation}. In the machine translation task, both HIT and MURIL perform significantly better than the other baselines. Moreover, HIT reports the best scores in BLEU and METEOR, whereas, MURIL yields better ROUGE score. On the other hand, in response generation, HIT outperforms MURIL in 3 out of 4 languages, with only Hindi being the exception, in which MURIL achieves $3.5\%$, $6\%$, and $5\%$ better scores in BLEU, ROUGE-L, and METEOR, respectively. 

Based on the observations made in the comparative study, we could conclude that HIT captures the semantics and syntax of texts with high code-mixed index (CMI)~\cite{gamback2014measuring} better than MURIL and ML-BERT, which are primarily pre-trained on monolingual corpus and work well on texts with low CMI. Even the other baselines -- HAN and CS-ELMo that utilize the hierarchical structure of code-mixed texts tend to outperform MURIL in classification tasks on texts having high code-mixing index.

\newcommand{\mlmtab}{




\begin{tabular}{|l|l|l|l|l|l|l}
\hline
\multicolumn{1}{|c|}{\multirow{2}{*}{Task}} & \multicolumn{1}{c|}{\multirow{2}{*}{MLM}} & \multicolumn{3}{c|}{Hindi}\\ \cline{3-5} 
\multicolumn{1}{|c|}{} & \multicolumn{1}{c|}{} & \multicolumn{1}{c}{Prec} & \multicolumn{1}{c}{Rec} & \multicolumn{1}{c|}{F1}\\ \hline \hline

\textbf{Humour} & w & 0.548 & 0.509 & 0.506\\ 
\cline{2-5}
& w/o & 0.592 & 0.596 & 0.593\\
\hline

\textbf{Sarcasm} & w & \bf 0.635 & \bf 0.573 & \bf 0.592\\ 
\cline{2-5}
& w/o & 0.478 & 0.487 & 0.475\\
\hline

\textbf{Sentiment} & w & 0.562 & 0.512 & 0.512\\ 
\cline{2-5}
& w/o & 0.745 & 0.702 & 0.703\\
\hline
\textbf{Intent} & w & \bf 0.931 & \bf 0.918 & \bf 0.922\\
\cline{2-5}
\textbf{detection} & w/o & 0.906 & 0.908 & 0.893\\
\hline
\end{tabular}%
\label{tab:mlm}
}

\newcommand{\zsltab}{
\begin{tabular}{|l|l|l|l|l|l|l}
\hline
\multicolumn{1}{|c|}{\multirow{2}{*}{Task}} & \multicolumn{1}{c|}{\multirow{2}{*}{ZSL}} & \multicolumn{3}{c|}{Hindi}\\ \cline{3-5} 
\multicolumn{1}{|c|}{} & \multicolumn{1}{c|}{} & \multicolumn{1}{c}{Prec} & \multicolumn{1}{c}{Rec} & \multicolumn{1}{c|}{F1}\\ \hline \hline

\textbf{Humour} & w & \bf 0.662 & \bf 0.668 & \bf 0.664\\ 
\cline{2-5}
& w/o & 0.592 & 0.596 & 0.593\\
\hline

\textbf{Sarcasm} & w & \bf 0.684 &  \bf 0.696 &  \bf 0.681\\ 
\cline{2-5}
& w/o & 0.478 & 0.487 & 0.475\\
\hline

\textbf{Sentiment} & w & \bf 0.832 &  \bf 0.794 &  \bf 0.796\\ 
\cline{2-5}
& w/o & 0.745 & 0.702 & 0.703\\
\hline
\textbf{Intent} & w & 0.892 & 0.889 & 0.882\\
\cline{2-5}
\textbf{detection} & w/o & 0.906 & 0.908 & 0.893\\

\hline

\end{tabular}%
\label{tab:zsl}
}

\newcommand{\hintra}{

\begin{tabular}{|c|c|l|l|l|l|l}
\hline
\multicolumn{1}{|c|}{\multirow{2}{*}{Source Task}} & \multicolumn{1}{c|}{\multirow{2}{*}{Fine-tune}} & \multicolumn{3}{c|}{Target Task}\\ \cline{3-5} 
\multicolumn{1}{|c|}{} & \multicolumn{1}{c|}{} & \multicolumn{1}{c}{POS} & \multicolumn{1}{c}{NER} & \multicolumn{1}{c|}{Sentiment}\\ \hline \hline

\multirow{2}{*}{\textbf{POS}} & w & \multirow{2}{*}{0.919} & 0.578 & 0.863\\ 
& w/o &  & 0.702 & \bf 0.890\\
\hline

\multirow{2}{*}{\textbf{NER}} & w & 0.873 & \multirow{2}{*}{0.745} & \bf 0.893\\ 
& w/o & \bf 0.924 & & \bf 0.885\\
\hline

\multirow{2}{*}{\textbf{Sentiment}} & w & \bf 0.928 & 0.691 & \multirow{2}{*}{0.703}\\ 
& w/o & \bf 0.936 & 0.729 &\\
\hline
\end{tabular}%
\label{tab:tl_hindi}
}

\newcommand{\spatra}{
\begin{tabular}{|c|c|l|l|l|l|l}
\hline
\multicolumn{1}{|c|}{\multirow{2}{*}{Source Task}} & \multicolumn{1}{c|}{\multirow{2}{*}{Fine-tune}} & \multicolumn{3}{c|}{Target Task}\\ \cline{3-5} 
\multicolumn{1}{|c|}{} & \multicolumn{1}{c|}{} & \multicolumn{1}{c}{POS} & \multicolumn{1}{c}{NER} & \multicolumn{1}{c|}{Sentiment}\\ \hline \hline

\multirow{2}{*}{\textbf{POS}} & w & \multirow{2}{*}{0.825} & 0.656 & 0.419\\ 
& w/o &  & \bf 0.710 & 0.417\\
\hline

\multirow{2}{*}{\textbf{NER}} & w & 0.663 & \multirow{2}{*}{0.684} & 0.446\\ 
& w/o & \bf 0.881 & & \bf 0.473\\
\hline

\multirow{2}{*}{\textbf{Sentiment}} & w & 0.732 & \bf 0.687 & \multirow{2}{*}{0.460}\\ 
& w/o & \bf 0.918 & \bf 0.969 &\\
\hline
\end{tabular}%
\label{tab:tl_sp}
}

\newcommand{\hispntra}{

\begin{tabular}{|c|c|l|l|l||l|l|l|}
\hline
\multicolumn{1}{|c|}{\multirow{3}{*}{Source Task}} & \multicolumn{1}{c|}{\multirow{3}{*}{Fine-tune}} & \multicolumn{6}{c|}{Target Task} \\ \cline{3-8} 
\multicolumn{1}{|c|}{} & \multicolumn{1}{c|}{} & \multicolumn{3}{c||}{Hindi-English} & \multicolumn{3}{c|}{Spanish-English}\\ \cline{3-8} 
\multicolumn{1}{|c|}{} & \multicolumn{1}{c|}{} & \multicolumn{1}{c}{POS} & \multicolumn{1}{c}{NER} & \multicolumn{1}{c||}{Sentiment} & \multicolumn{1}{c}{POS} & \multicolumn{1}{c}{NER} & \multicolumn{1}{c|}{Sentiment}\\ \hline \hline

\multirow{2}{*}{\textbf{POS}} & w & \multirow{2}{*}{0.919} & 0.578 & \bf 0.863 & \multirow{2}{*}{0.825} & 0.656 & 0.419\\ 
& w/o &  & 0.702 & \bf 0.890 & & \bf 0.710 & 0.417\\
\hline

\multirow{2}{*}{\textbf{NER}} & w & 0.873 & \multirow{2}{*}{0.745} & \bf 0.893 & 0.663 & \multirow{2}{*}{0.684} & 0.446\\ 
& w/o & \bf 0.924 & & \bf 0.885 & \bf 0.881 & & \bf 0.473\\
\hline

\multirow{2}{*}{\textbf{Sentiment}} & w & \bf 0.928 & 0.691 & \multirow{2}{*}{0.703} & 0.732 & \bf 0.687 & \multirow{2}{*}{0.460}\\ 
& w/o & \bf 0.936 & 0.729 & & \bf 0.918 & \bf 0.969 &\\
\hline
\end{tabular}%
\label{tab:tl_hindi_spanish}
}

\begin{table*}[t]
  \centering
  \subfloat[Masked Language Modeling]{\scalebox{0.65}\mlmtab}%
  \quad
  \subfloat[Zero-Shot Learning]{\scalebox{0.65}\zsltab}
  \quad
  \subfloat[Transfer Learning for Hi-En and Sp-En]{\scalebox{0.65}\hispntra}
  \caption{Experimental results of our HIT model on (a) Masked Language Modeling for Hindi-English, (b) Zero-Shot Learning for Hindi-English, and (c) Transfer learning for Hindi-English and Spanish-English datasets. For MLM and ZSL, we highlight the rows where the model achieves better result with pre-training. For transfer learning, we highlight the rows where the model achieves better performance by transferring knowledge from the source task to target task.}
  \label{tab:mlmzsl}%
  \vspace{-2.5mm}
\end{table*}

\section{Generalization through Pre-training}
In order to make our framework more robust and task-invariant, we adopt several pre-training strategies to learn a task-invariant code-mixed representation from texts. We consolidate the language-specific datasets and conduct  pre-training. With this strategy, our representation learning model leverages a larger dataset to learn a generic representation for each text that can be utilized in any downstream supervised task. Precisely, we adopt masked language modeling (MLM), zero-shot learning (ZSL), and transfer learning. Among these, only MLM is a pre-training objective, while the other two are learning paradigms that utilize the knowledge either from other datasets, or from an already trained model. With these pre-training objectives, the high-level learning of the model is shared across tasks, thus making our representation learning task-invariant and generalized. 
The pre-trained semantic knowledge can be utilized by adding a separate task-specific dense layer during the fine-tuning stage. This way we can ensure that HIT does not overfit on any particular task, rather, learns the underlying semantics of the code-mixed texts. 

In the subsequent sections, we demonstrate several analyses to showcase the task-invariance and generalizability aspects of HIT representation, learned through pre-training. 

\subsection{Pre-training Objectives}
\textbf{Masked Language Modeling (MLM): } We model MLM following Devlin {\em et al.} \cite{devlin_bert_2019} to robustly learn semantics and contextual representations that are task-invariant. Given a sentence, we choose 15\% of the tokens for modifying under this objective as follows - (a) 80\% of the chosen words are replaced with mask token ``[MASK]", (b) 10\% of the chosen tokens are replaced with a random token from the vocabulary, and (c) 10\% of the tokens are retained without any replacement. This forms the input, and the unmodified sentence is the output to be generated. We implement HIT for extracting character, subword, and word embeddings coupled with FAME to pre-train this objective. For evaluating on downstream tasks, we implement a simple feed-forward network classifier with HIT as its backbone embeddings to compare the performance with and without MLM pre-training.

\textbf{Zero-Shot Learning (ZSL): } In this approach, we leverage representation learning across the input text and the target classes. Given a set of input texts \{$a_1, a_2, \dots, a_n$\} and target classes \{$c^{(j)}_1, c^{(j)}_2, \dots, c^{(j)}_n$\} for each task $j$, where each input text belongs to one target class for each task, we prepare the input dataset for each input text $a_i$ as follows -- (a) An input pair ($a_i, c^{(j)}_i)$ where $c^{(j)}_i$ is the true target class and (b) An input pair ($a_i, \overline{c}^{(j)}_i)$ where $\overline{c}^{(j)}_i$ is any target class except $c^{(j)}_i$. We employ a negative sampling to generate the false target class randomly. With this dataset, we process the HIT representations for both input text and target class and compute cosine similarity to classify the instance as entailment or contradiction. During inference, ZSL objective helps us in achieving a robust semantic representation for a text without explicitly using any text label.

Due to the limited data availability, we only use Hindi-English code-mixed texts for the zero-shot training. Also, for simplicity, we use only sequence classification tasks -- sentiment, humour, and sarcasm classification for training this objective. These tasks are developed with a similar semantic objective, which makes them easier to bind together in a multi-task learning framework. A total of 7 labels are used to train the ZSL objective, namely - \emph{humour, non-humour, sarcasm, non-sarcasm, positive sentiment, negative sentiment, neutral sentiment}. Intent detection task dataset is used only for testing. 
\subsection{Results with Pre-training Objectives}
{\bf Masked Language Modeling:} We show results for both with (w) and without (w/o) MLM pre-training in Table \ref{tab:mlm}. Due to the availability of the Hindi dataset across 4 tasks, we only conduct this analysis on Hindi language. We consolidate the available datasets containing Hindi scripts for the pre-training purpose. The representation learned through MLM is used and subsequently fine-tuned in sequence classification tasks. We observe a decrease in humour and sentiment tasks when we adapt the initial embedding from the pre-trained model. On the other hand, with MLM pre-training, we achieve $12\%$ better F1-score in sarcasm detection and $3\%$ better F1-score in intent classification. Moreover, we observe that HIT even outperforms MURIL in the intent classification task. 

{\bf Zero-Shot Learning:} We show the results in Table \ref{tab:zsl}. Comparatively, zero-shot learning has performed considerably better as compared to vanilla HIT (without any pre-training objective). On the sentiment classification task, F1-score of $0.796$ is achieved -- a significant $9\%$ jump over the vanilla HIT model. Further, we observe similar phenomena on the sarcasm detection ($0.681$ compared to $0.475$) and humour classification ($0.664$ compared to $0.593$) tasks. Moreover, ZSL outperforms on all the tasks; hence, asserting the scope of pre-training objectives that can leverage larger training corpus. Furthermore, in comparison with MURIL, HIT with zero-shot learning achieves better results on the humour and sarcasm classification tasks; thus demonstrating the necessity of proper pre-training objective to learn better semantics.

{\bf Transfer Learning:} To completely capture different setups of learning across code-mixed datasets, we explore a straightforward transfer learning setup with (w) and without (w/o) fine-tuning. As with previous learning setups, these experiments shed light on the model's capabilities to learn linguistic and semantic features rather than task-specific features. For brevity, we choose Hindi and Spanish datasets, as they have multiple tasks in them and maximum overlap in terms of tasks. Therefore, we select the PoS, NER, and sentiment classification tasks for comparison. Table \ref{tab:tl_hindi_spanish} presents results for Hindi-English and Spanish-English code-mixed languages, respectively. For each case, we train our HIT model on one source task, and subsequently run experiments on the other two target tasks. For the Hindi code-mixed dataset, except for NER as the source task, we observe positive performance transfer in other cases. Considering PoS as the source task, we observe an improvement in sentiment classification as target. Similarly, with sentiment as the source task, we observe improvements in both PoS and NER tasks. Likewise in the Spanish dataset, HIT reports improvements in NER with PoS as the source task. We observe similar phenomena with NER and sentiment as source tasks. 
This shows that our HIT model is able to generalise and learn linguistic and semantic representation given sufficient number of diverse training set.

\newcommand{\senterr}{
\begin{tabular}{|c|c|c|c|}
\hline
Labels & Pos & Neg & Neu\\
\hline
Pos & \textit{{\color{blue}0.85}} & 0.05 & \bf 0.10\\
\hline
Neg & 0.03 &  \textit{{\color{blue}0.87}} & \bf 0.10\\
\hline
Neu & - & 0.02 & \textit{{\color{blue}0.98}}\\
\hline

\end{tabular}%
\label{tab:sentiment_confusion}
}

\newcommand{\nererr}{

\begin{tabular}{|c|c:c|c:c|c:c|c|}
\hline
Tags & B-Per & I-Per & B-Loc & I-Loc & B-Org & I-Org & O\\
\hline
B-Per & \textit{{\color{blue}0.85}} & 0.01 & - & - & - & - & \bf 0.14\\
\hdashline
I-Per & 0.03 & \textit{{\color{blue}0.87}} & - & - & - & - & \bf 0.10\\
\hline
B-Loc & - & - & \textit{{\color{blue}0.92}} & - & 0.02 & 0.02 & 0.05\\
\hdashline
I-Loc & - & 0.08 & - & \textit{\color{blue}0.85} & - & - & 0.07\\
\hline
B-Org & 0.02 & 0.02 & - & - & \textit{\color{blue}0.82} & - & \bf 0.14\\ \hdashline
I-Org & - & - & - & - & - & \textit{\color{blue}0.71} & \bf 0.29\\
\hline
O & - & - & - & - & \bf 0.47 & - & \textit{\color{blue}0.53}\\

\hline

\end{tabular}%
\label{tab:ner_confusion}
}

\newcommand{\sarerr}{
\begin{tabular}{|c|c|c|c|}
\hline
Labels & Sarcasm & Non-Sarcasm\\
\hline
Sar & \textit{{\color{blue}0.67}} & 0.33\\
\hline
Non-Sar & 0.05 &  \textit{{\color{blue}0.95}}\\
\hline
\end{tabular}%
\label{tab:sarcasm_confusion}
}

\newcommand{\humerr}{
\begin{tabular}{|c|c|c|c|}
\hline
Labels & Humour & Non-Humour\\
\hline
Hum & \textit{{\color{blue}0.63}} & 0.37\\
\hline
Non-Hum & 0.21 &  \textit{{\color{blue}0.79}}\\
\hline
\end{tabular}%
\label{tab:humour_confusion}
}

\newcommand{\inthindi}{

\begin{tabular}{|c|c|c|c|c|c|c|c|c|c|c|c|c|c|c|c|c|c|}
\hline
Intent & Ack & Affirm & Cant.help & Confirm & Deny & Expl-conf & Impl-conf & Inform & Negate & Offer & Repeat & Requalts & Reqmore & Request & Silence & Thankyou & Welcomemsg \\
\hline
Ack & \textit{\color{blue}0.962} & - & - & - & - & - & - & 0.02 & - & - & - & - & - & 0.018 & - & - & -\\
\hline
Affirm & - & \textit{\color{blue}0.877} & 0.015 & - & - & - & - & - & - & - & - & \textbf{0.108} & - & - & - & - & -\\
\hline
Cant.help & - & - & \textit{\color{blue}0.997} & - & - & - & - & - & 0.002 & - & - & - & - & - & 0.001 & - & - \\
\hline
Confirm & - & - & - & \textit{\color{blue}0.687} & - & - & - & \textbf{0.297} & - & - & - & 0.016 & - & - & - & - & - \\
\hline
Deny & - & - & - & - & \textit{\color{blue}0.909} & - & - & 0.091 & - & - & - & - & - & - & - & - & - \\
\hline
Expl-conf & - & - & - & 0.002 & - & \textit{\color{blue}0.989} & - & 0.009 & - & - & - & - & - & - & - & - & -\\
\hline
Impl-conf & - & - & \textbf{0.239} & - & - & - & \textit{\color{blue}0.474} & \textbf{0.279} & - & - & - & - & - & 0.008 & - & - & -\\
\hline
Inform & - & - & - & - & - & 0.018 & - & \textit{\color{blue}0.979} & 0.001 & 0.001 & 0.001 & - & - & - & - & - & -\\
\hline
Negate & - & 0.024 & - & - & - & 0.024 & - & \textbf{0.280} & \textit{\color{blue}0.512} & - & 0.085 & 0.012 & - & 0.048 & - & 0.015 & -\\
\hline
Offer & - & - & - & - & - & - & 0.017 & - & - & \textit{\color{blue}0.979} & 0.004 & - & - & - & - & - & -\\
\hline
Repeat & - & - & - & - & - & - & - & \textbf{0.698} & - & - & \textit{\color{blue}0.262} & 0.040 & - & - & - & - & -\\
\hline
Requalts & - & - & - & - & - & - & - & 0.090 & 0.020 & - & 0.017 & \textit{\color{blue}0.873} & - & - & - & - & -\\
\hline
Reqmore & - & - & - & - & - & - & - & - & - & - & - & - & \textit{\color{blue}1.00} & - & - & - & -\\
\hline
Request & 0.009 & 0.005 & - & - & - & - & - & 0.039 & - & - & - & - & - & \textit{\color{blue}0.947} & - & - & -\\
\hline
Silence & 0.055 & 0.035 & - & - & - & - & - & 0.096 & - & - & - & - & - & - & \textit{\color{blue}0.814} & - & -\\
\hline
Thankyou & - & - & - & - & - & - & - & - & - & - & - & - & - & - & - & \textit{\color{blue}1.00} & -\\
\hline
Welcomemsg & - & - & - & - & - & - & - & - & - & - & - & - & - & - & - & - & \textit{\color{blue}1.00}\\
\hline

\hline

\end{tabular}%
\label{tab:intent_hindi}
}

\newcommand{\slothindi}{

\begin{tabular}{|c|c:c|c:c|c:c|c:c|c:c|c|c:c|c:c|}
\hline
Slot & B-Food & I-Food & B-Phone & I-Phone & B-Addr & I-Addr & B-Name & I-Name & B-Postcode & I-Postcode & B-Pricerange & I-Pricerange & B-Slot & I-Slot & O\\
\hline
B-Food & \textit{\color{blue}0.962} & - & 0.004 & - & - & - & - & - & - & - & - & - & 0.004 &  - & 0.030\\
\hdashline
I-Food & - & \textit{\color{blue}0.877} & - & - & 0.013 & - & - & - & - & - & - & - &  - &  - & \textbf{0.110}\\
\hline
B-Phone & - & - & \textit{\color{blue}0.996} & - & - & - & - & - & - & - & 0.001 & - &  - &  - & 0.003\\
\hdashline
I-Phone & - & - & - & \textit{\color{blue}0.990} & - & - & - & 0.001 & - & - & - & - &  - &  - & 0.009\\
\hline
B-Addr & - & - & - & - & \textit{\color{blue}0.777} & - & - & - & - & - & - &  - &  - & - & \textbf{0.223}\\
\hdashline
I-Addr & - & - & - & - & - & \textit{\color{blue}0.820} & - & 0.001 & - & - & 0.018 &  - & - & - &  \textbf{0.161}\\
\hline
B-Name & - & 0.020 & 0.010 & - & - & - & \textit{\color{blue}0.678} & - & 0.010 & - & 0.010 & - &  0.010 & - &  \textbf{0.262}\\
\hdashline
I-Name  & - & - & - & - & - & - & - & \textit{\color{blue}0.990} & - & - & - & - &  - &  - & 0.010\\
\hline
B-Postcode & - & - & - & - & - & - & - & - & \textit{\color{blue}1.00} & - & - & - &  - &  - & -\\
\hdashline
I-Postcode & - & - & - & - & - & - & - & - & - & \textit{\color{blue}1.00} & - & - &  - &  - & -\\
\hdashline
B-Pricerange & - & - & - & 0.002 & - & - & - & 0.024 & - & - & \textit{\color{blue}0.834} & - &  - &  - & 0.140\\
\hline
I-Pricerange & - & - & - & - & - & - & - & - & - & - & - & - &  - &  - & -\\
\hline
B-Slot & - & - & - & - & - & - & - & 0.021 & 0.002 & - & - & - &  \textit{\color{blue}0.667} - &  & \textbf{0.310}\\
\hline
I-Slot & - & - & - & - & - & - & - & - & - & - & - & - &  - &  - & -\\
\hline
O & - & - & 0.01 & 0.021 & - & 0.012 & 0.024 & - & 0.033 & - & - & - &  - &  - & \textit{\color{blue}0.900}\\
\hline

\end{tabular}%
\label{tab:slot_hindi}
}

\begin{table*}%
  \centering
  \subfloat[NER]{\scalebox{0.6}\nererr}
  \qquad
  \subfloat[Sentiment]{\scalebox{0.7}\senterr}%
  \qquad
  \subfloat[Sarcasm]{\scalebox{0.8}\sarerr}
  \qquad
  \subfloat[Humour]{\scalebox{0.8}\humerr}
  
  \subfloat[Intent detection]{\scalebox{0.71}\inthindi} \qquad
  \subfloat[Slot filling]{\scalebox{0.71}\slothindi}
  \caption{Confusion matrices in percentage (\%) on Hindi code-mixed datasets.}%
  \label{tab:confusion}%
\end{table*}



\section{Error Analyses and Discussion}
In this section, we elaborately perform quantitative and qualitative analyses of HIT and the effectiveness of pre-training objectives for improving downstream task performance for code-mixed texts.

We present confusion matrices of HIT for all 6 code-mixed Hinglish classification tasks in Table \ref{tab:confusion}. We observe that class-wise true positives in respective tasks are significant in majority of the cases. Also, except for a few cases, false-negative and false-positives are non-significant. In NER (c.f. Table \ref{tab:ner_confusion}), we observe that HIT majorly faces problem with the other (`O') class -- false negatives for the `person' and `organization' tags are relatively higher. Moreover, we also observe a significant 47\% false positive rate for `B-Org' and `O'. In comparison, we observe 20\% false positives for the `neural' class in sentiment classification. On the other hand, we observe an overall higher error rates for the humour (Table \ref{tab:humour_confusion}) and sarcasm (Table \ref{tab:sarcasm_confusion}) classification tasks -- true positives are relatively lower than other tasks and false negative are also higher. This could be due to the complex nature of the humour and sarcasm detection tasks. In contrast, HIT performs significantly well for the intent classification (Table \ref{tab:intent_hindi}) and slot-filling (Table \ref{tab:slot_hindi}) tasks -- both false-positives and false-negatives are minute in comparison with the true positives. The most misclassified label in intent detection is \emph{inform} whose utterances encompass longer sentences with characterstics of multiple intent values.

\begin{table*}[ht!]
    \centering
    \resizebox{\textwidth}{!}{
    \begin{tabular}{|l|l|p{30em}|p{30em}|}
         \hline
         $1$ & I/p & \multicolumn{2}{l|}{** START\_CONVERSATION **} \\  \cdashline{2-4}
         & Res & {\bf Ref:} Hello , Cambridge restaurant system mein aapka swagat hai . Aap pa sakte hai restaurant by area , price range ya food type . Aap ko kaise help kar sakta hu main ? (\textit{Hello, welcome to the Cambridge restaurant system. You can request restaurant by area, price range, or food type. How can I help you?}) & {\bf HIT:} Hello , Cambridge restaurant system mein aapka swagat hai . Aap pa sakte hai restaurant by area , price range ya food type . Aap ko kaise help kar sakta hu main ? (\textit{Hello, welcome to the Cambridge restaurant system. You can request restaurant by area, price range, or food type. How can I help you?})\\\hline

         $2$ & I/p & \multicolumn{2}{l|}{** START\_CONVERSATION **} \\  \cdashline{2-4}
         & Res & {\bf Ref:} Aap kaunse type ka food lene chahege? (\textit{Which food type would you prefer?}) & {\bf HIT:} \textcolor{red}{Helloap} kaunse type ka food lene chahege? (\textit{Helloap which food type would you prefer?}) \\ \hline
         
         $3$ & I/p & \multicolumn{2}{l|}{North part of town meh persian restaurant hai (\textit{Is there any persian restaurant in the north part of town?})} \\ \cdashline{2-4}
         & Res & {\bf Ref:} \textcolor{blue}{Sorry} but yahan koi bhi restaurant \textcolor{blue}{Persian} khana serve nahin karta (\textit{Sorry but no restaurant serves Persian food here}) & {\bf HIT:} \textcolor{red}{Hello} but yahan koi bhi restaurant \textcolor{red}{Malaysian} khana serve nahin karta (\textit{Hello but no restaurant serves Malaysian food  here}) \\\hline
     \end{tabular}}
    \caption{Actual and generated responses with input text are shown for a single round of user-bot interaction. \textcolor{red}{Red} color denotes contextually incorrect responses. The input represented here is only the current user utterance for which the bot response is predicted. To understand the complete model input at each instance, refer to Section \ref{sec:data_info} A.}
    \label{tab:error:response}
\end{table*}

Furthermore, we conduct fine-grained analyses on response generation task and compare the responses generated by HIT and the ground-truth values in Table~\ref{tab:error:response}. Examples show generated responses as bot's utterances. 
The first response in the table is a standard greeting that is correctly generated by our model. In the second response, the model generated the sentence with a single error only -- it commits error on the first word. However, the semantic of the generated response is arguably intact. On the other hand, in the third example, the model fails to capture the essential user request considering the type of cuisine requested -- it predicts \emph{Malaysian} instead of \emph{Persian}. Similar to the previous case, the model generates the reminder of the response correctly.

{\bf Masked Language Modeling:} 
To understand the robustness and effectiveness of the MLM representation learning, we extract HIT's embedding with and without pre-trained MLM backend and conduct a fine-grained analysis. We extract the spectral embedding~\cite{belkin2003laplacian} of these embeddings for two tasks -- sentiment classification and sarcasm classification for Hindi-English texts. We observe that the pretraining version of MLM helps our model in achieving a cohesive representation for semantically similar texts. Further, to understand how well these embeddings are clustered, we perform $k$-means clustering on the HIT embeddings with the number of clusters being matched with the number of classes in the respective tasks ($n\_clusters=3$ and $2$ for the sentiment and sarcasm classification tasks, respectively). We utilize Silhoutte score~\cite{rousseeuw1987silhouettes} (higher the better) and Davies-Bouldin (DB) index~\cite{davies1979cluster} (lower the better) to indicate the robustness of learned representations irrespective of the downstream tasks. For sentiment classification, HIT with pre-trained MLM achieves a Silhoutte score of 0.536, as compared to HIT without MLM pre-training that achieves 0.379. Similarly, HIT with MLM achieves a DB index of 0.612 against 0.997 achieved by HIT without MLM. We observe similar phenomenon for the sarcasm classification task as well -- HIT with MLM reports better Silhoutte ($+0.262$) and DB scores ($-0.585$), compared to the counterpart model without pre-trained MLM backend. This quantitative analysis demonstrates the strength of MLM pre-training for achieving semantically richer representation, even without proper supervision.

{\bf Zero-Shot Learning:} 
We also analyze the effectiveness of HIT in zero-shot learning setup. As discussed earlier, our approach to ZSL leverages shared information among three tasks viz. \textit{sentiment classification}, \textit{humour classification}, and \textit{sarcasm detection} -- a total of 7 labels (\emph{humour, non-humour, sarcasm, non-sarcasm, positive, negative, neutral}) are used to learn the ZSL objective. One of the foremost observations in this setup is the association among the three labels -- \emph{non-humour, non-sarcasm}, and \emph{neutral sentiment}. The average Pearson's correlation among these three classes, based on the HIT prediction probability, turns out to be $0.74$ with p-value $\leq 0.01$. Moreover, we observe that HIT learns to group these inputs close together leveraging the similarities between them. On the other hand, sarcastic and humorous texts majorly consist of exaggerated terms to invoke emphasis to the context and correlate with the `\textit{negative}' sentiment. This adds to the fundamental robustness of ZSL approach wherein the model can group text representations in the embedding space such that different tasks are seamlessly learnt in a shared space aiding one another. 

\section{Conclusion and Future Work}
\label{sec:con}
In this paper, we presented an extensive research on code-mixed dataset on several tasks. We explored a novel attention mechanism, FAME (Fused attention mechanism) that significantly outperforms other code-mixed and multilingual representation learning methods. We conducted extensive experiments on sentiment classification, named-entity recognition, parts-of-speech tagging, machine translation, sarcasm detection, and humour classification. Further on conversational datasets, we explored response generation, intent detection, and slot filling tasks. Finally, to emphasize the generalizability of our model, we observed model's performance on Masked Language Modeling and zero-shot learning pre-training objectives. We showed that our HIT model is robust in learning contextual representations across Indian languages (and Spanish) and we put forward the model to be reused by the research community. We argue that this would open a new avenue in utilizing attention-based models in analyzing low-resource languages. An interesting direction for the future can be to utilize these knowledge in understanding how code-mixed languages are generated, and adapted in conversational settings like - chatbots, social medias, etc.

{\small
\bibliography{assets/styles/code_mix}

\begin{thebibliography}{10}
\providecommand{\url}[1]{#1}
\csname url@rmstyle\endcsname
\providecommand{\newblock}{\relax}
\providecommand{\bibinfo}[2]{#2}
\providecommand\BIBentrySTDinterwordspacing{\spaceskip=0pt\relax}
\providecommand\BIBentryALTinterwordstretchfactor{4}
\providecommand\BIBentryALTinterwordspacing{\spaceskip=\fontdimen2\font plus
\BIBentryALTinterwordstretchfactor\fontdimen3\font minus
  \fontdimen4\font\relax}
\providecommand\BIBforeignlanguage[2]{{%
\expandafter\ifx\csname l@#1\endcsname\relax
\typeout{** WARNING: IEEEtran.bst: No hyphenation pattern has been}%
\typeout{** loaded for the language `#1'. Using the pattern for}%
\typeout{** the default language instead.}%
\else
\language=\csname l@#1\endcsname
\fi
#2}}

\bibitem{PARSHAD2016375}
\BIBentryALTinterwordspacing
R.~D. Parshad, S.~Bhowmick, V.~Chand, N.~Kumari, and N.~Sinha, ``What is india
  speaking? exploring the “hinglish” invasion,'' \emph{Physica A:
  Statistical Mechanics and its Applications}, vol. 449, pp. 375--389, 2016.
  [Online]. Available:
  \url{https://www.sciencedirect.com/science/article/pii/S0378437116000236}
\BIBentrySTDinterwordspacing

\bibitem{pratapa_word_2018}
A.~Pratapa, M.~Choudhury, and S.~Sitaram, ``Word embeddings for code-mixed
  language processing,'' in \emph{Proceedings of the 2018 conference on
  empirical methods in natural language processing}, 2018, pp. 3067--3072.

\bibitem{aguilar_english_2020}
G.~Aguilar and T.~Solorio, ``From english to code-switching: Transfer learning
  with strong morphological clues,'' \emph{arXiv preprint arXiv:1909.05158},
  2019.

\bibitem{sengupta_hit_2021}
A.~Sengupta, S.~K. Bhattacharjee, T.~Chakraborty, and M.~S. Akhtar, ``Hit-a
  hierarchically fused deep attention network for robust code-mixed language
  representation,'' in \emph{Findings of the Association for Computational
  Linguistics: ACL-IJCNLP 2021}, 2021, pp. 4625--4639.

\bibitem{le_self-attentive_2020}
H.~Le, T.~Tran, and S.~Venkatesh, ``Self-attentive associative memory,'' in
  \emph{International Conference on Machine Learning}.\hskip 1em plus 0.5em
  minus 0.4em\relax PMLR, 2020, pp. 5682--5691.

\bibitem{vaswani_attention_nodate}
A.~Vaswani, N.~Shazeer, N.~Parmar, J.~Uszkoreit, L.~Jones, A.~N. Gomez,
  {\L}.~Kaiser, and I.~Polosukhin, ``Attention is all you need,'' in
  \emph{NIPS}, 2017, pp. 5998--6008.

\bibitem{upadhyay_cross-lingual_2016}
S.~Upadhyay, M.~Faruqui, C.~Dyer, and D.~Roth, ``Cross-lingual models of word
  embeddings: An empirical comparison,'' \emph{arXiv preprint
  arXiv:1604.00425}, 2016.

\bibitem{ruder_survey_2019}
S.~Ruder, I.~Vuli{\'c}, and A.~S{\o}gaard, ``A survey of cross-lingual word
  embedding models,'' \emph{Journal of Artificial Intelligence Research},
  vol.~65, pp. 569--631, 2019.

\bibitem{akhtar_solving_2018}
M.~S. Akhtar, P.~Sawant, S.~Sen, A.~Ekbal, and P.~Bhattacharyya, ``Solving data
  sparsity for aspect based sentiment analysis using cross-linguality and
  multi-linguality.''\hskip 1em plus 0.5em minus 0.4em\relax Association for
  Computational Linguistics, 2018.

\bibitem{faruqui_improving_2014}
M.~Faruqui and C.~Dyer, ``Improving vector space word representations using
  multilingual correlation,'' in \emph{Proceedings of the 14th Conference of
  the European Chapter of the Association for Computational Linguistics}, 2014,
  pp. 462--471.

\bibitem{hermann_multilingual_2014}
K.~M. Hermann and P.~Blunsom, ``Multilingual models for compositional
  distributed semantics,'' \emph{arXiv preprint arXiv:1404.4641}, 2014.

\bibitem{luong_bilingual_2015}
M.-T. Luong, H.~Pham, and C.~D. Manning, ``Bilingual word representations with
  monolingual quality in mind,'' in \emph{Proceedings of the 1st Workshop on
  Vector Space Modeling for Natural Language Processing}, 2015, pp. 151--159.

\bibitem{labutov_generating_2014}
I.~Labutov and H.~Lipson, ``Generating code-switched text for lexical
  learning,'' in \emph{Proceedings of the 52nd Annual Meeting of the
  Association for Computational Linguistics (Volume 1: Long Papers)}, 2014, pp.
  562--571.

\bibitem{gupta_uncovering_2018}
D.~Gupta, P.~Lenka, A.~Ekbal, and P.~Bhattacharyya, ``Uncovering code-mixed
  challenges: A framework for linguistically driven question generation and
  neural based question answering,'' in \emph{Proceedings of the 22nd
  Conference on Computational Natural Language Learning}, 2018, pp. 119--130.

\bibitem{banerjee_dataset_2018}
S.~Banerjee, N.~Moghe, S.~Arora, and M.~M. Khapra, ``A dataset for building
  code-mixed goal oriented conversation systems,'' in \emph{COLING}, 2018, pp.
  3766--3780.

\bibitem{srivastava_phinc_2020}
V.~Srivastava and M.~Singh, ``Phinc: A parallel hinglish social media
  code-mixed corpus for machine translation,'' \emph{arXiv preprint
  arXiv:2004.09447}, 2020.

\bibitem{prabhu_towards_2016}
A.~Joshi, A.~Prabhu, M.~Shrivastava, and V.~Varma, ``Towards sub-word level
  compositions for sentiment analysis of hindi-english code mixed text,'' in
  \emph{COLING}, 2016, pp. 2482--2491.

\bibitem{yang_hierarchical_2016}
Z.~Yang, D.~Yang, C.~Dyer, X.~He, A.~Smola, and E.~Hovy, ``Hierarchical
  attention networks for document classification,'' in \emph{Proceedings of the
  2016 conference of the North American chapter of the association for
  computational linguistics: human language technologies}, 2016, pp.
  1480--1489.

\bibitem{devlin_bert_2019}
J.~Devlin, M.-W. Chang, K.~Lee, and K.~Toutanova, ``Bert: Pre-training of deep
  bidirectional transformers for language understanding,'' \emph{arXiv preprint
  arXiv:1810.04805}, 2018.

\bibitem{khanuja2021muril}
S.~Khanuja, D.~Bansal, S.~Mehtani, S.~Khosla, A.~Dey, B.~Gopalan, D.~K. Margam,
  P.~Aggarwal, R.~T. Nagipogu, S.~Dave, \emph{et~al.}, ``Muril: Multilingual
  representations for indian languages,'' \emph{arXiv preprint
  arXiv:2103.10730}, 2021.

\bibitem{pratapa2018language}
A.~Pratapa, G.~Bhat, M.~Choudhury, S.~Sitaram, S.~Dandapat, and K.~Bali,
  ``Language modeling for code-mixing: The role of linguistic theory based
  synthetic data,'' in \emph{ACL-2018}, pp. 1543--1553.

\bibitem{brown2020language}
T.~B. Brown, B.~Mann, N.~Ryder, M.~Subbiah, J.~Kaplan, P.~Dhariwal,
  A.~Neelakantan, P.~Shyam, G.~Sastry, A.~Askell, \emph{et~al.}, ``Language
  models are few-shot learners,'' \emph{arXiv preprint arXiv:2005.14165}, 2020.

\bibitem{gupta2021unsupervised}
A.~Gupta, S.~Menghani, S.~K. Rallabandi, and A.~W. Black, ``Unsupervised
  self-training for sentiment analysis of code-switched data,'' \emph{arXiv
  preprint arXiv:2103.14797}, 2021.

\bibitem{yadav2021zera}
S.~Yadav and T.~Chakraborty, ``Zera-shot sentiment analysis for code-mixed
  data,'' in \emph{AAAI-2021}, vol.~35, no.~18, 2021, pp. 15\,941--15\,942.

\bibitem{ba2016layer}
J.~L. Ba, J.~R. Kiros, and G.~E. Hinton, ``Layer normalization,'' \emph{arXiv
  preprint arXiv:1607.06450}, 2016.

\bibitem{bansal_code-switching_2020}
S.~Bansal, V.~Garimella, A.~Suhane, J.~Patro, and A.~Mukherjee,
  ``Code-switching patterns can be an effective route to improve performance of
  downstream nlp applications: A case study of humour, sarcasm and hate speech
  detection,'' \emph{arXiv preprint arXiv:2005.02295}, 2020.

\bibitem{chakravarthi_corpus_nodate}
B.~R. Chakravarthi, V.~Muralidaran, R.~Priyadharshini, and J.~P. McCrae,
  ``Corpus creation for sentiment analysis in code-mixed tamil-english text,''
  in \emph{Proceedings of SLTU and CCURL}, 2020, pp. 202--210.

\bibitem{patwa_semeval-2020_2020}
P.~Patwa, G.~Aguilar, S.~Kar, S.~Pandey, S.~Pykl, B.~Gamb{\"a}ck,
  T.~Chakraborty, T.~Solorio, and A.~Das, ``Semeval-2020 task 9: Overview of
  sentiment analysis of code-mixed tweets,'' in \emph{Proceedings of the
  Fourteenth Workshop on Semantic Evaluation}, 2020, pp. 774--790.

\bibitem{singh2018named}
V.~Singh, D.~Vijay, S.~S. Akhtar, and M.~Shrivastava, ``Named entity
  recognition for hindi-english code-mixed social media text,'' in \emph{Proc.
  7th Named Entities Workshop}, pp. 27--35.

\bibitem{aguilar-etal-2018-named}
\BIBentryALTinterwordspacing
G.~Aguilar, F.~AlGhamdi, V.~Soto, M.~Diab, J.~Hirschberg, and T.~Solorio,
  ``Named entity recognition on code-switched data: Overview of the {CALCS}
  2018 shared task,'' in \emph{CALCS}, 2018, pp. 138--147. [Online]. Available:
  \url{https://www.aclweb.org/anthology/W18-3219}
\BIBentrySTDinterwordspacing

\bibitem{singh-etal-2018-twitter}
\BIBentryALTinterwordspacing
K.~Singh, I.~Sen, and P.~Kumaraguru, ``A {T}witter corpus for {H}indi-{E}nglish
  code mixed {POS} tagging,'' in \emph{Proc. 6th Int. Workshop on NLP for
  Social Media}.\hskip 1em plus 0.5em minus 0.4em\relax Melbourne, Australia:
  ACL, July 2018, pp. 12--17. [Online]. Available:
  \url{https://www.aclweb.org/anthology/W18-3503}
\BIBentrySTDinterwordspacing

\bibitem{alghamdi-etal-2016-part}
\BIBentryALTinterwordspacing
F.~AlGhamdi, G.~Molina, M.~Diab, T.~Solorio, A.~Hawwari, V.~Soto, and
  J.~Hirschberg, ``Part of speech tagging for code switched data,'' in
  \emph{Proc. 2nd Workshop on Computational Approaches to Code Switching},
  2016, pp. 98--107. [Online]. Available:
  \url{https://www.aclweb.org/anthology/W16-5812}
\BIBentrySTDinterwordspacing

\bibitem{gupta-etal-2020-semi}
\BIBentryALTinterwordspacing
D.~Gupta, A.~Ekbal, and P.~Bhattacharyya, ``A semi-supervised approach to
  generate the code-mixed text using pre-trained encoder and transfer
  learning,'' in \emph{Findings of the ACL: EMNLP 2020}, 2020, pp. 2267--2280.
  [Online]. Available:
  \url{https://www.aclweb.org/anthology/2020.findings-emnlp.206}
\BIBentrySTDinterwordspacing

\bibitem{henderson2014second}
M.~Henderson, B.~Thomson, and J.~D. Williams, ``The second dialog state
  tracking challenge,'' in \emph{SIGDIAL-2014}, pp. 263--272.

\bibitem{bedi_multi-modal_2021}
M.~Bedi, S.~Kumar, M.~S. Akhtar, and T.~Chakraborty, ``Multi-modal sarcasm
  detection and humor classification in code-mixed conversations,'' \emph{IEEE
  Transactions on Affective Computing}, 2021.

\bibitem{lin_rouge_nodate}
C.-Y. Lin, ``Rouge: A package for automatic evaluation of summaries,'' in
  \emph{Text summarization branches out}, 2004, pp. 74--81.

\bibitem{papineni_bleu_2001}
K.~Papineni, S.~Roukos, T.~Ward, and W.-J. Zhu, ``Bleu: a method for automatic
  evaluation of machine translation,'' in \emph{Proceedings of the 40th annual
  meeting of the Association for Computational Linguistics}, 2002, pp.
  311--318.

\bibitem{banerjee_meteor_2005}
S.~Banerjee and A.~Lavie, ``Meteor: An automatic metric for mt evaluation with
  improved correlation with human judgments,'' in \emph{Proceedings of the acl
  workshop on intrinsic and extrinsic evaluation measures for machine
  translation and/or summarization}, 2005, pp. 65--72.

\bibitem{hochreiter_long_1997}
S.~Hochreiter and J.~Schmidhuber, ``Long short-term memory,'' \emph{Neural
  computation}, vol.~9, no.~8, pp. 1735--1780, 1997.

\bibitem{peters_deep_2018}
\BIBentryALTinterwordspacing
M.~E. Peters, M.~Neumann, M.~Iyyer, M.~Gardner, C.~Clark, K.~Lee, and
  L.~Zettlemoyer, ``Deep contextualized word representations,'' in
  \emph{Proceedings of the 2018 Conference of the North {A}merican Chapter of
  the Association for Computational Linguistics: Human Language Technologies,
  Volume 1 (Long Papers)}.\hskip 1em plus 0.5em minus 0.4em\relax New Orleans,
  Louisiana: Association for Computational Linguistics, June 2018, pp.
  2227--2237. [Online]. Available: \url{https://aclanthology.org/N18-1202}
\BIBentrySTDinterwordspacing

\bibitem{hu2020xtreme}
J.~Hu, S.~Ruder, A.~Siddhant, G.~Neubig, O.~Firat, and M.~Johnson, ``Xtreme: A
  massively multilingual multi-task benchmark for evaluating cross-lingual
  generalisation,'' in \emph{ICML}, 2020, pp. 4411--4421.

\bibitem{kingma_adam_2017}
D.~P. Kingma and J.~Ba, ``Adam: A method for stochastic optimization,''
  \emph{arXiv preprint arXiv:1412.6980}, 2014.

\bibitem{sutskever_sequence_nodate}
I.~Sutskever, O.~Vinyals, and Q.~V. Le, ``Sequence to sequence learning with
  neural networks,'' in \emph{NIPS}, 2014, pp. 3104--3112.

\bibitem{bahdanau_neural_2016}
D.~Bahdanau, K.~Cho, and Y.~Bengio, ``Neural machine translation by jointly
  learning to align and translate,'' \emph{arXiv preprint arXiv:1409.0473},
  2014.

\bibitem{see_get_2017}
A.~See, P.~J. Liu, and C.~D. Manning, ``Get to the point: Summarization with
  pointer-generator networks,'' \emph{arXiv preprint arXiv:1704.04368}, 2017.

\bibitem{gamback2014measuring}
B.~Gamb{\"a}ck and A.~Das, ``On measuring the complexity of code-mixing,'' in
  \emph{ICON}, 2014, pp. 1--7.

\bibitem{belkin2003laplacian}
M.~Belkin and P.~Niyogi, ``Laplacian eigenmaps for dimensionality reduction and
  data representation,'' \emph{Neural computation}, vol.~15, no.~6, pp.
  1373--1396, 2003.

\bibitem{rousseeuw1987silhouettes}
P.~J. Rousseeuw, ``Silhouettes: a graphical aid to the interpretation and
  validation of cluster analysis,'' \emph{Journal of computational and applied
  mathematics}, vol.~20, pp. 53--65, 1987.

\bibitem{davies1979cluster}
D.~L. Davies and D.~W. Bouldin, ``A cluster separation measure,'' \emph{IEEE
  TPAMI}, no.~2, pp. 224--227, 1979.

\end{thebibliography}
\bibliographystyle{assets/styles/IEEEtran}}

\end{document}